\documentclass{article}

\usepackage{wrapfig}
\usepackage{CJKutf8} 
\usepackage{tabularx} 
\usepackage{longcat_style}

\usepackage[utf8]{inputenc} 
\usepackage[T1]{fontenc}    
\usepackage{newunicodechar}
\newunicodechar{，}{,}
\usepackage{hyperref}       
\usepackage{xcolor}
\usepackage{svg} 
\hypersetup{
    colorlinks=true,      
    linkcolor=blue,      
    urlcolor=blue,       
    citecolor=blue,      
    linkbordercolor=blue, 
    urlbordercolor=blue,
    citebordercolor=blue,
    pdfborderstyle={/S/U/W 1}, 
}

\usepackage{xurl}            
\usepackage{booktabs}       
\usepackage{amsfonts}       
\usepackage{nicefrac}       
\usepackage{microtype}      
\usepackage{lipsum}		
\usepackage{graphicx}
\usepackage{natbib}
\usepackage{doi}
\usepackage{amsmath}
\usepackage{amssymb} 
\usepackage{xspace}
\usepackage{enumitem}
\usepackage{multirow}
\usepackage{subcaption} 
\usepackage{makecell}
\usepackage{hyperref, cleveref}
\setlist[itemize]{leftmargin=*}
\setlist[enumerate]{leftmargin=*}
\setlist[description]{leftmargin=*}
\newcommand{\longcatthink}{LongCat-Flash-Thinking\xspace}

\newcommand{\longcatbase}{LongCat-Flash-Base\xspace}

\newcommand{\longcatchat}{LongCat-Flash-Chat\xspace}

\usepackage{colortbl}
\definecolor{mygray}{gray}{.88}
\definecolor{mycyan}{cmyk}{.15,0,0,0}
\definecolor{mycyan2}{cmyk}{.85,0,0,0}

\definecolor{mygreen}{rgb}{0.19, 0.79, 0.02}
\definecolor{midnightgreen}{rgb}{0.0, 0.29, 0.33}

\title{Introducing \longcatthink: A Technical Report}

\author{ Meituan LongCat Team \\
	\texttt{longcat-team@meituan.com} \\
}

\renewcommand{\headeright}{\raisebox{-0.2\height}{\includegraphics[height=1.8em]{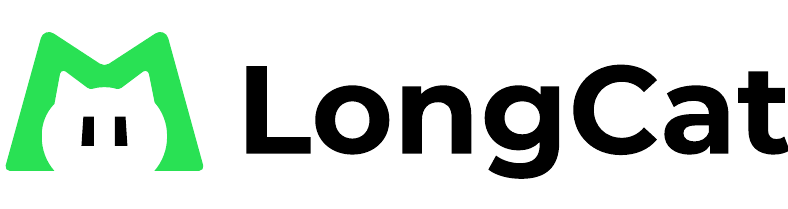}}}
\renewcommand{\shorttitle}{Introducing \longcatthink: A Technical Report}

\begin{document}
\maketitle

\begin{abstract}
We present~{\longcatthink}, an efficient $560$-billion-parameter open-source Mixture-of-Experts (MoE) reasoning model.
Its advanced capabilities are cultivated through a meticulously crafted training process, beginning with long Chain-of-Thought (CoT) data cold-start and culminating in large-scale Reinforcement Learning (RL). We first employ a well-designed cold-start training strategy, which significantly enhances the reasoning potential and equips the model with specialized skills in both formal and agentic reasoning.
Then, a core innovation is our domain-parallel training scheme, which decouples optimization across distinct domains (e.g., STEM, Code, Agentic) and subsequently fuses the resulting expert models into a single, nearly Pareto-optimal model.
This entire process is powered by our Dynamic ORchestration for Asynchronous rollout (DORA) system, a large-scale RL framework that delivers a greater than threefold training speedup over synchronous methods on tens of thousands of accelerators. 
As a result, {\longcatthink} achieves state-of-the-art performance among open-source models on a suite of complex reasoning tasks.
The model exhibits exceptional efficiency in agentic reasoning, reducing average token consumption by $64.5\%$ (from $19,653$ to $6,965$) on AIME-25, without degrading task accuracy.
We release {\longcatthink} to promote further advances in reasoning systems and agentic AI research.

\textbf{LongCat Chat}: \href{https://longcat.ai}{https://longcat.ai} \\
\textbf{Huggingface}: \href{https://huggingface.co/meituan-longcat/LongCat-Flash-Thinking}{https://huggingface.co/meituan-longcat/LongCat-Flash-Thinking}\\
\textbf{Github}: \href{https://github.com/meituan-longcat/LongCat-Flash-Thinking}{https://github.com/meituan-longcat/LongCat-Flash-Thinking}\\
\end{abstract}

\begin{figure}[hb]
\centerline{\includegraphics[width=\linewidth]{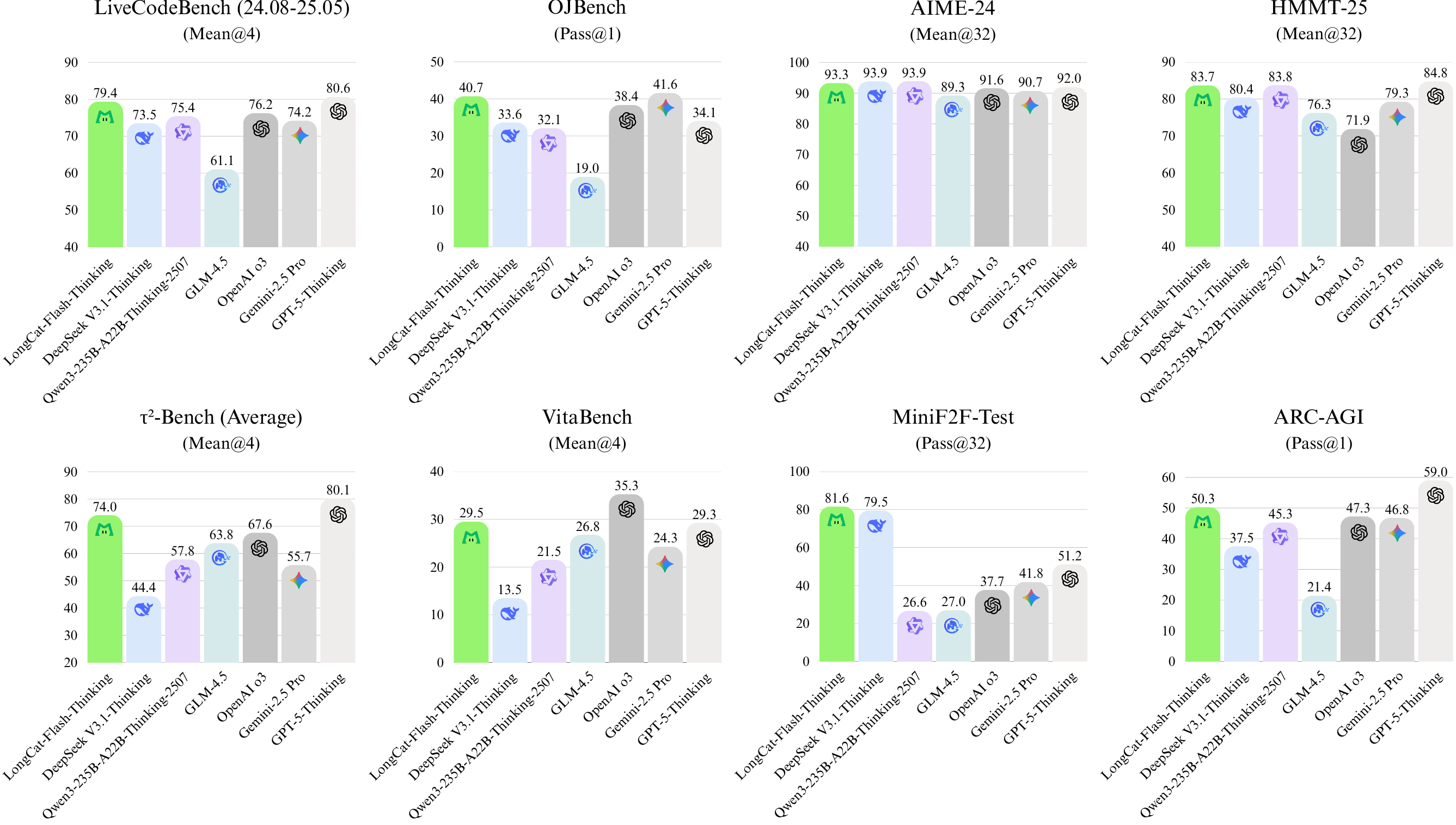}}
\caption{\small
The comparison of average performance on reasoning benchmarks. The four models on the left are open-weight LLMs, while the others are closed-weight LLMs. 
}
\vspace{-.3cm}
\label{fig:qualitative_example}
\end{figure}


\newpage

\section{Introduction}
In recent years, the frontier of Large Language Models (LLMs) has shifted decisively toward enhancing their reasoning capabilities, pushing the boundaries of Artificial General Intelligence (AGI). 
State-of-the-art models such as OpenAI's o1~\citep{openai_o1}, OpenAI's o3 \citep{o3}, Gemini 2.5 \citep{comanici2025gemini}, DeepSeek-R1 \citep{guo2025deepseek}, Qwen3 \citep{yang2025qwen3}, and GLM-4.5 \citep{5team2025glm45agenticreasoningcoding} showcase this trend, demonstrating profound abilities in complex logic, mathematics, code generation, and agentic tasks. This advancement is largely driven by a new paradigm: leveraging large-scale Reinforcement Learning (RL)~\citep{sutton1998reinforcement} not only to refine models but also to enable deeper, more extensive reasoning at inference time. By dynamically allocating more computational FLOPs to extend Chain-of-Thought (CoT)~\citep{wei2022chain}, specialized models like OpenAI's o1 and Gemini 2.5 Pro have achieved breakthrough performance on formidable challenges, including olympiad-level mathematics, competitive programming, and complex agentic scenarios.

\begin{figure}[h]
\centerline{\includegraphics[width=\linewidth]{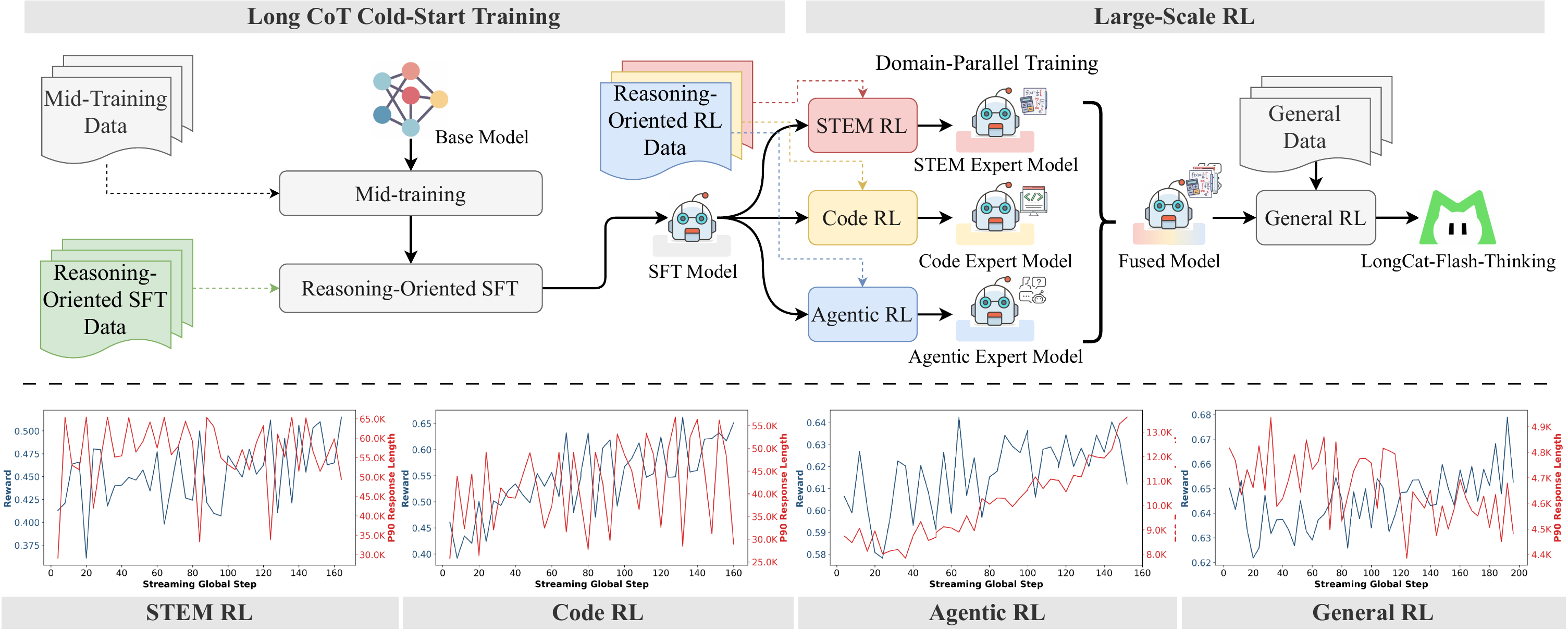}}
\caption{\small
The training pipeline of~{\longcatthink}. We first perform mid-training and SFT to cultivate the base model’s foundational reasoning abilities. Then, we introduce a domain-parallel training scheme during the large-scale RL phase to obtain multiple domain-specific experts.
Finally, a general RL stage followed by expert models fusion to improve the general ability.
}
\label{fig:parallel_training_pipeline}
\end{figure}

In this work, we introduce {\longcatthink}, an efficient open-source Mixture-of-Experts (MoE) reasoning model that establishes a new state-of-the-art for open-source reasoning models. Built upon our foundational {\longcatbase} model~\citep{meituan2025longcat_flash_chat}, {\longcatthink} ($560$B total parameters: 27B active on average) excels on complex logic, STEM, coding, and agent tasks. The development of {\longcatthink} follows a meticulously designed two-phase pipeline: Long CoT Cold-Start Training and Large-Scale RL. In the first phase, we aim to build the model's foundational reasoning abilities. This begins with a curriculum learning strategy during mid-training to strengthen intrinsic capabilities, followed by a targeted Supervised Fine-Tuning (SFT) stage on reasoning-intensive and agentic data to prepare the model for advanced learning. The second phase scales up this potential through an efficient RL framework, built upon our Dynamic ORchestration for Asynchronous rollout (DORA) system for industrial-scale asynchronous training. To address the stability challenges in asynchronous RL training, we adapt and extend the Group Relative Policy Optimization (GRPO)~\citep{grpo} algorithm for a robust exploration-exploitation balance. A key innovation in this phase is our domain-parallel training scheme, which simultaneously optimizes the model across distinct domains and subsequently merges the resulting domain-expert models into a fused model~\citep{yu2024language}. A final general RL stage further refines the fused model, improving its robustness, safety, and human alignment for real-world applications. The overview of the training pipeline is illustrated in Figure~\ref{fig:parallel_training_pipeline}.

Our work presents three core contributions:
\begin{itemize}
    \item \textbf{Domain-Parallel RL Training and Fusion Methodology}: To overcome the instability of traditional mixed-domain RL training, we design a domain-parallel scheme that decouples optimization across STEM, coding, and agentic tasks. This approach not only stabilizes training but also allows us to fuse the resulting domain-expert models into a nearly Pareto-optimal final model that excels across all specialties.
    \item \textbf{Pioneering Industrial-Scale RL Infrastructure}:  Our DORA system provides the robust backbone for RL training. Its asynchronous architecture delivers a greater than threefold speedup over synchronous frameworks, enabling stable training on tens of thousands of accelerators. This industrial-grade system supported a massive RL investment of nearly $20\%$ of pre-training compute, making our advanced methodology feasible at scale. We also highlight its novel features, including elastic colocation and efficient KV-cache reuse, which are detailed in Section~\ref{subsec:rl_infra}.
    \item \textbf{Broad and Efficient Advanced Reasoning}: We significantly extend the model's capabilities into challenging domains, achieving exceptional performance and efficiency. To improve agentic capabilities, we propose a dual-path reasoning approach to select high-value training queries that benefit most from tool integration. We complement this with an automated pipeline to construct high-quality, tool-augmented reasoning trajectories for training. For agentic tool use, {\longcatthink} leads to a $64.5\%$ reduction in token consumption on the AIME-25 benchmark while preserving accuracy. 
    For formal reasoning, we developed an expert-iteration pipeline integrated with a Lean4 server to synthetically generate verified proofs, systematically instilling a capability absent in most LLMs.
\end{itemize}
These contributions culminate in a model that not only achieves state-of-the-art performance on a diverse set of benchmarks (Figure~\ref{fig:qualitative_example}), but also establishes a clear advantage in the crucial domains of formal proving and agentic reasoning over its open-source peers.

\section{Long CoT Cold-Start Training}
\begin{figure}[t]
\centerline{\includegraphics[width=\linewidth]{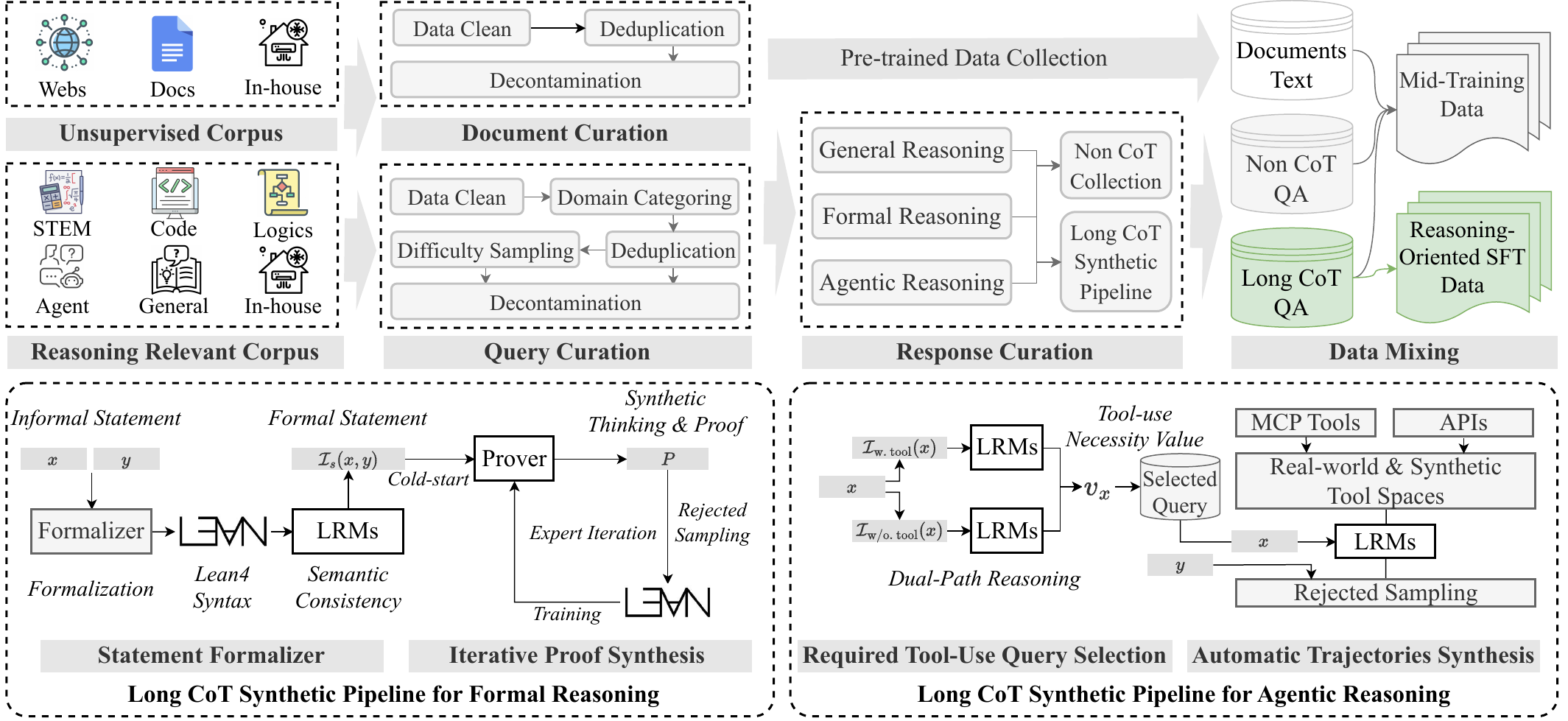}}
\caption{\small
The data curation pipeline for cold-start training.
}
\label{fig:cold_start_data_pipeline}
\end{figure}

We conduct long CoT cold-start training and large-scale RL based on the~{\longcatbase} model \citep{meituan2025longcat_flash_chat}, equipping it with advanced reasoning capabilities.
Benefit from the zero-computation experts~\citep{jin2024moe++} and shortcut-connected MoE structure~\citep{cai2024shortcut}, {\longcatthink} achieves large efficiency advantages compared to models of comparable performance.
In this section, we concentrate on augmenting the long CoT reasoning capability~\citep{wei2022chain} of our base model through a multi-stage curriculum learning approach. We first introduce a mid-training phase, during which the base model learns from diverse reasoning data to enhance its \textit{fundamental reasoning capabilities} and \textit{potential for RL}. Subsequently, we incorporate a compact SFT phase. In addition to high-quality general reasoning data, we specifically expose the model to \textit{formal reasoning} and \textit{agentic reasoning} capabilities, both of which are designed to effectively enhance the reasoning performance.
The overview of the cold-start data curation pipeline is shown in Figure~\ref{fig:cold_start_data_pipeline}.

\subsection{Mid-training: Reasoning Capability Enhancement}

While our foundational pre-training yields a model with robust general capabilities~\citep{meituan2025longcat_flash_chat}, we identified a critical limitation on its ability to handle complex reasoning tasks. Although fine-tuning base models followed by RL training has substantially improved the reasoning performance, we observe that these models often produce homogeneous reasoning patterns, which hinder their ability to reflect deeply and obtain correct solutions for challenging problems.

This deficiency is twofold, stemming from the composition of general pre-training corpora. 
First, while vast, these corpora are heavily weighted toward general text, resulting in an insufficient proportion of data from reasoning-intensive domains such as STEM and coding. 
Second, and more critically, explicit long CoT patterns, which form the very structure of methodical reasoning, are naturally scarce even within that specialized data. 
This dual-faceted data gap stunts the model’s intrinsic reasoning potential, creating a significant bottleneck for subsequent fine-tuning stages. 

To overcome this, and inspired by analyses of reasoning boundaries in Large Reasoning Models (LRMs)~\citep{yue2025does}, our approach transforms the standard mid-training phase~\citep{meituan2025longcat_flash_chat} into a carefully balanced curriculum. 
The objective is to cultivate the model's latent reasoning abilities (effectively "cold-starting" them) without degrading its foundational generalist knowledge, thereby setting a stronger starting point for the subsequent long CoT SFT.

\noindent\paragraph{Training Recipe}
Our curriculum is built upon a meticulously curated dataset of reasoning-intensive problems across STEM and coding domains. 
The STEM collection features a diverse range of mathematics, physics, and chemistry problems sourced from academic archives, textbooks, and proprietary data, with a strong emphasis on competition-level challenges. 
Our curation process prioritizes the problems requiring multi-step logical reasoning over those solvable by simple fact retrieval.
For algorithmic programming tasks, we aggregated problems from diverse open-source code competition datasets.
This curated data is then strategically infused into the training corpus. 
We employ a rigorous quality-control pipeline, using a hybrid of heuristic rules and LLM-as-a-Judge methods for filtering, deduplication, and decontamination.
Crucially, we carefully adjust the data mixing ratio, balancing the reasoning-intensive data against the original mid-training data. 
This ensures the model develops foundational reasoning skills without degrading its generalist capabilities.
The detailed data curation and mixing are provided in Appendix~\ref{appendix:rbe}.

\noindent\paragraph{Evaluation}
Prior to the formal training on the LongCat-Flash-Base~\citep{meituan2025longcat_flash_chat}, we first conducted a preliminary experiment to validate the effectiveness of our reasoning-enhancement mid-training. 
This pilot study was performed on a small-scale, in-house MoE model that shares an identical architecture.
We employ a repeated sampling strategy with the $\mathit{pass@k}$~\citep{yue2025does, brown2024large} metric to evaluate the model's reasoning capability.
To ensure an unbiased estimation of this metric, we follow the methodology proposed by~\citet{chen2021evaluating}.
Formally, given one query $x$ from a query set $\mathcal{D}$, let the model be $\pi_{\theta}$, where $\theta$ denotes to the parameter, we generate $N$ ($N > 0$) responses as $\{y_i\}_{i=1}^{N}$, where $y_i=\pi_{\theta}(x)$ represents one response.
Hence, the $\mathit{pass@k}$ is defined as:
\begin{equation}
    pass@k := \mathbb{E}_{x_i\sim\mathcal{D}}\bigg[1 - \frac{{N - C_i \choose k}}{{N \choose k}}\bigg],
\end{equation}
where $C_i$ ($C_i \leq N$) denotes the number of correct answers.

\begin{figure}[t]
\centerline{\includegraphics[width=.95\linewidth]{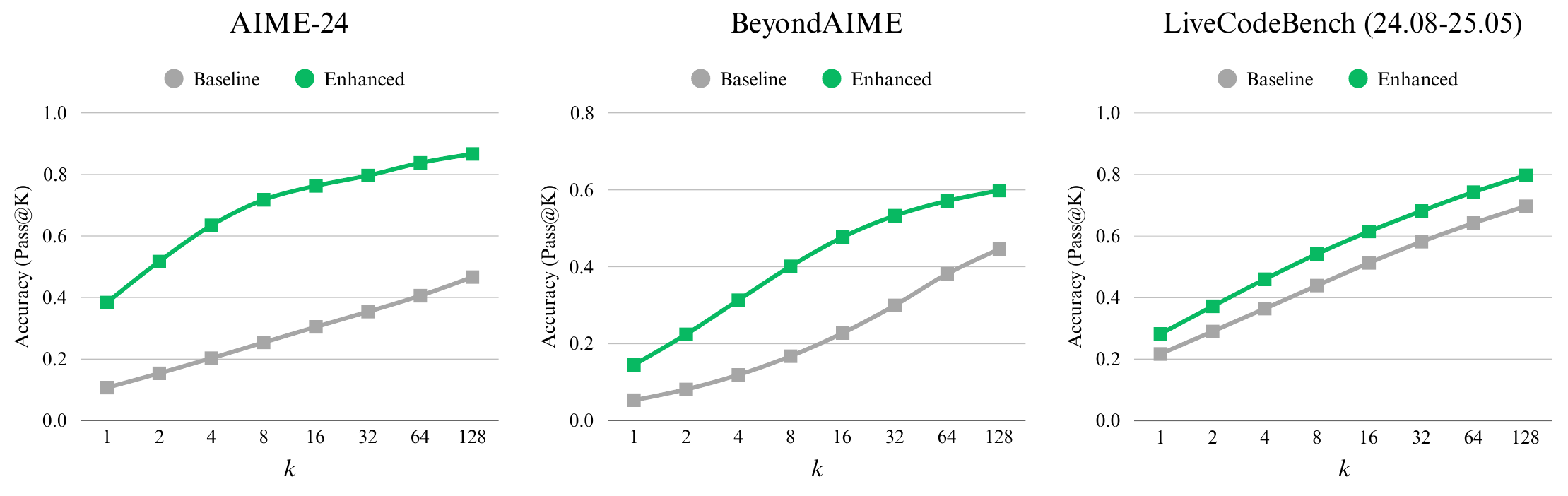}}
\caption{\small
The comparison of reasoning capability ($\mathit{pass@k}$). ``Baseline'' represents our in-house small base model, ``Enhanced'' represents the modified small base model after reasoning capability enhancement in the mid-training phase. 
}
\label{fig:base_model_passk}
\end{figure}

Figure~\ref{fig:base_model_passk} illustrates our evaluation results on three benchmarks: AIME-24, BeyondAIME, and LiveCodeBench (LCB) (24.08-25.05).
The results reveal a clear trend: a higher proportion of reasoning-intensive data in mid-training consistently enhances the model's reasoning performance across all metrics, from $\mathit{pass@}\text{1}$ to $\mathit{pass@}\text{128}$. 
This effect is significant across all sampling complexities, with $\mathit{pass@}\text{1}$ scores improving by $27.7\%$ on AIME-24, $9.3\%$ on BeyondAIME, and $6.5\%$ on LCB.
Notably, the improvements are even more substantial for higher k-values like $\mathit{pass@}\text{64}$ and $\mathit{pass@}\text{128}$, demonstrating that this approach effectively broadens the model's reasoning boundary. 
These findings led us to integrate this strategy into our LongCat-Flash-Thinking mid-training process.

\subsection{Reasoning-oriented SFT}
Following the mid-training curriculum, we introduce a reasoning-oriented SFT stage to align the model with high-quality instruction-following patterns and enhance its specialized reasoning capabilities, thereby establishing a strong foundation for subsequent large-scale RL. In addition to general reasoning, we focus on advancing~{\longcatthink} on formal reasoning and agentic reasoning, which can cultivate the model's reasoning ability with formal language and real-world tools, respectively.

\subsubsection{General Reasoning}
To enhance general reasoning capabilities, we curate a diverse, high-quality training dataset from multiple domains: STEM, code, logic, and general QA. The construction process involves a rigorous pipeline for both prompt curation and response generation, as illustrated in the following. Further details on the data processing for each domain are provided in Appendix~\ref{appendix:sft}.

First, for prompt curation, we implement a multi-stage filtering process. 
1) Initial Screening: We use an LLM-as-Judge~\citep{zhou2025libra} approach to eliminate low-quality or unanswerable queries, such as incomplete statements. For code-related data, we select problems that feature a clear description, a robust set of at least five unit tests, and an executable judging script. 
2) Ground-Truth Validation: To verify correctness, a model-based voting mechanism is employed. 
This involves automatically generating diverse responses to identify and filter out prompts with inconsistent or potentially erroneous ground truths. 
3) Difficulty Filtering: Except for general QA, we estimate problem difficulty via the pass rate from expert models. 
Prompts with pass rates above a certain threshold are discarded as too simple. 
The final prompt set is then sampled from the filtered pool according to the difficulty distribution.

Second, for response generation, we employ a rejection sampling methodology. 
Candidate responses are synthesized for each prompt, with LongCat-Flash-Chat~\citep{meituan2025longcat_flash_chat} serving as the primary generator. 
These candidates are then evaluated through a combination of rule-based and model-based judgments to select the highest-quality response for our final training data.

\subsubsection{Formal Reasoning}
The recent success of models like Qwen2.5-Math~\citep{yang2024qwen2}, Kimina-Prover~\citep{wang2025kimina}, DeepSeek-Prover~\citep{Xin2025DeepSeek, Ren2025DeepSeek}, and Seed-Prover~\citep{chen2025seedproverdeepbroadreasoning} highlights the immense potential of LRMs to accelerate research in formal reasoning tasks like Automatic Theorem Proving (ATP). 
To help realize this potential and empower researchers, we introduce significant enhancements to our model's formal reasoning capabilities. 
Our work aims to provide a robust and versatile foundation upon which the community can build and explore new scientific frontiers.
To achieve this, we focus on ATP, a representative and challenging task in formal reasoning. 
We introduce a novel methodology to systematically enhance our model's capabilities in this domain.
The pipeline is shown in the lower left corner of Figure~\ref{fig:cold_start_data_pipeline}.




\noindent\paragraph{Task Definition}
Formally, the task of ATP is to generate a valid proof $\mathcal{P}$ for a given formal statement.
The process starts with an informal problem, consisting of a natural language question $x$ and its answer $y$.
This is the first converted into a formal statement $s = \mathcal{I}_s(x,y)$ by an autoformalizer $\mathcal{I}_s$.
The model $\pi_{\theta}$ then generates a proof candidate $\mathcal{P} = \pi_{\theta}(s)$. 
A verifier $\mathcal{V}$ checks the proof, yielding a binary outcome $\mathcal{V}(\mathcal{P}, s) \in \{\text{PASS}, \text{FAIL}\}$.
Our work focuses on whole-proof generation, where the entire proof is produced in a single turn from the formal statement.


\noindent\paragraph{Statement Formalization}
We collect multiple competition-grade mathematical problems 
and perform data deduplication and decontamination.
Since the original data only contains natural language problems, we train an 8B-based autoformalizer model to translate each informal statement (containing the original question and answer) into a formal statement.
We then perform a two-stage filtering process to ensure its correctness:
1) Syntax Filtering: We follow~\citet{wang2025kimina}'s work to develop the Lean4 Server~\footnote{\url{https://github.com/project-numina/kimina-lean-server}.} (v4.15). Each generated formal statement is concatenated with the placeholder ``:= by sorry'', and compiled through the Lean4 Server. Thus, the statements with syntax errors are removed.
2) Semantic Filtering: We found that autoformalization can occasionally alter the original problem's meaning.
To address this, we employ a model-based semantic filter to identify and discard formal statements that are inconsistent with their informal counterparts.



\noindent\paragraph{Iterative Proof Synthesis}
Our proof synthesis follows an iterative data enhancement strategy, beginning with a cold-start process and progressively refined through expert iteration.
For this purpose, we utilize our reasoning-enhanced LongCat-Flash-Base model as the foundation for the prover, which is systematically improved throughout this process.
The iteration pipeline is: 
\begin{itemize}
    \item \textbf{Cold-start Prover Training:} The goal of this phase is to build an initial dataset to train the baseline prover. First, to identify a set of solvable problems, we filter our formal statements by leveraging existing theorem-proving tools. Statements that can be successfully verified are retained, forming our initial set of (statement, proof) pairs. Next, to enrich this data with reasoning steps, we employ a model-based synthesis approach to generate a natural language "thinking process" for each pair. This creates the final training triples of (statement, thinking process, proof), which are then used to perform an initial SFT on our LongCat-Flash-Base model.
    

    \item \textbf{Expert Iteration:} This phase iteratively expands the dataset and enhances the prover. In each iteration: 1) The current prover attempts to generate proofs for all formal statements that remain unsolved. 2) Newly generated proofs are verified, and the successful (statement, proof) pairs are added to our dataset. 3) These new pairs are then enriched with a synthesized thinking process using the same model-based approach. 4) Finally, we aggregate all curated training triples and retrain the prover from scratch. This self-improvement loop repeats for a fixed number of iterations.
    
\end{itemize}



Through this iterative process, we curated a substantial corpus of high-quality training instances, each containing a formal statement, a synthesized thinking process, and a verified proof. 
This dataset was then used to comprehensively enhance the formal theorem-proving capabilities of our~{\longcatthink}.


\subsubsection{Agentic Reasoning}


Agentic reasoning can be embodied with tools use, interpreting, and complex problem solving~\citep{zeng2025glm, openai_o1}.
Existing datasets are often plagued by instances where the model can produce satisfactory answers without actually invoking the tool. 
Such data provide limited utility for real-world agentic behaviors, as they lack the challenge of leveraging external tools for problem-solving.
To alleviate this issue, we focus on identifying and retaining high-quality queries that genuinely require tool assistance, thereby fostering the development of robust agentic abilities. 

\noindent\paragraph{Required Tool-Use Query Selection}

To curate a dataset of queries that genuinely require tool use, we begin by aggregating candidates from diverse sources, including open-source datasets (e.g., ToolBench~\citep{xu2023tool}, ToolLLM~\citep{qin2023toolllm}) and in-house data, performing standard deduplication and decontamination. 
We then introduce a novel dual-path evaluation pipeline to assess the ``tool necessity'' for each query. 
Specifically, for a given query $x\in\mathcal{D}$, we prompt a baseline model to generate $N$ solution trajectories under two distinct settings: one with access to tools $(\mathcal{I}_{\text{w. tool}})$ and one without $(\mathcal{I}_{\text{w/o. tool}})$. 
This yields two sets of responses:
\begin{equation}
    \mathcal{Y}_{\text{w/. tool}}=\{y_i|y_i=\pi_{\theta}(\mathcal{I}_{\text{w/. tool}}(x))\}_{i=1}^{N}
    \text{, and }
    \mathcal{Y}_{\text{w/o. tool}}=\{y_i|y_i=\pi_{\theta}(\mathcal{I}_{\text{w/o. tool}}(x))\}_{i=1}^{N}\text{.}
\end{equation}
Next, these responses are evaluated by an LLM-as-a-Judge to calculate the pass rates $s_{\text{w/. tool}}(x)$ and $s_{\text{w/o. tool}}(x)$ for $\mathcal{Y}_{\text{w/. tool}}$ and $\mathcal{Y}_{\text{w/o. tool}}$, respectively.
The tool-necessity value $v_x$ is then defined as the performance gain from using tools: $v_x = s_{\text{w/. tool}}(x) - s_{\text{w/o. tool}}(x)$.
A higher $v_x$ signifies that a query is difficult to solve with internal knowledge alone but becomes manageable with tool assistance.
Suppose that $\tau_1$, $\tau_2$, $\tau_3$ are the pre-defined thresholds, we select queries based on a set of thresholds: $\{x|v_x>\tau_1\land s_{\text{w/. tool}}(x)>\tau_2\land s_{\text{w/o. tool}}(x)<\tau_3, x\in\mathcal{D}\}$, ensuring that our final dataset consists of problems where tools are not just helpful, but indispensable.
\noindent\paragraph{Automatic Trajectories Synthesis}
After selecting high-value queries, we synthesize corresponding high-quality solution trajectories. 
To support a wide range of tasks, we first build a versatile environment with diverse tool APIs, including MCP servers and simulated tools for both single and multi-turn interactions.
For each selected query, we employ a powerful generative model to produce multiple candidate trajectories, spanning from simple tool calls to complex, multi-step workflows. 
These candidates are then rigorously evaluated by a panel of model-based judges on criteria such as correctness, logical consistency, and completeness of tool use.
Only trajectories that pass this evaluation are retained.
The validated trajectories are then standardized into a consistent format, ensuring logical completeness and clarity in reasoning steps.
Finally, we stratify these trajectories by complexity, based on factors such as the number of tool calls (single vs. multi-turn), dependency structures (sequential vs. parallel), and reasoning depth (e.g., lookup, multi-hop, planning), to facilitate curriculum-based learning and targeted model enhancements.

\subsubsection{Training Recipe}
For the SFT stage, we employ a sophisticated data curation strategy to balance the diverse and complex scenarios from our three reasoning-oriented datasets. 
This strategy included strict data decontamination protocols to ensure zero exposure to test data during training. 
To further bolster general reasoning, we up-sample data from the STEM and coding domains. 
In addition, we curate the final training instances based on several response behavior features we defined, such as average response length, reflection density, and query clustering. 
The objective of this approach is to markedly enhance performance on broad reasoning tasks while preserving proficiency in specialized areas like agentic tool use and formal proof generation. 
The final data mixing proportions are detailed in Figure~\ref{fig:sft_ratio}.

The SFT is performed on our reasoning-enhanced base model from the mid-training phase. 
We utilize the AdamW optimizer with a learning rate of $3e\text{-}5$, and train the model for $2$ epochs. 
To accommodate complex and extended reasoning chains, we set the context length to $48$K tokens.



\section{Large-Scale Reinforcement Learning}
RL is a critical phase for advancing the reasoning capabilities of LLMs, offering superior token efficiency and generalization over SFT.
However, applying RL to LLMs is notoriously challenging. The training process is often unstable, highly sensitive to hyperparameters, and incurs substantial system overhead that complicates industrial-scale deployment.
To overcome these hurdles, we develop a comprehensive, three-pronged solution: 1) At the system level, we build DORA, a robust distributed RL framework that supports asynchronous training and flexible accelerator usage to ensure stability and efficiency. 2) At the algorithm level, we introduce several modifications to stabilize training and enhance adaptability. 3) At the reward level, we design a versatile reward system capable of handling both verifiable and non-verifiable tasks, ensuring broad domain applicability.
The following subsections will detail our RL infrastructure, algorithmic enhancements, and reward design.

\begin{figure}
    \centering
    \includegraphics[width=0.9 \linewidth]{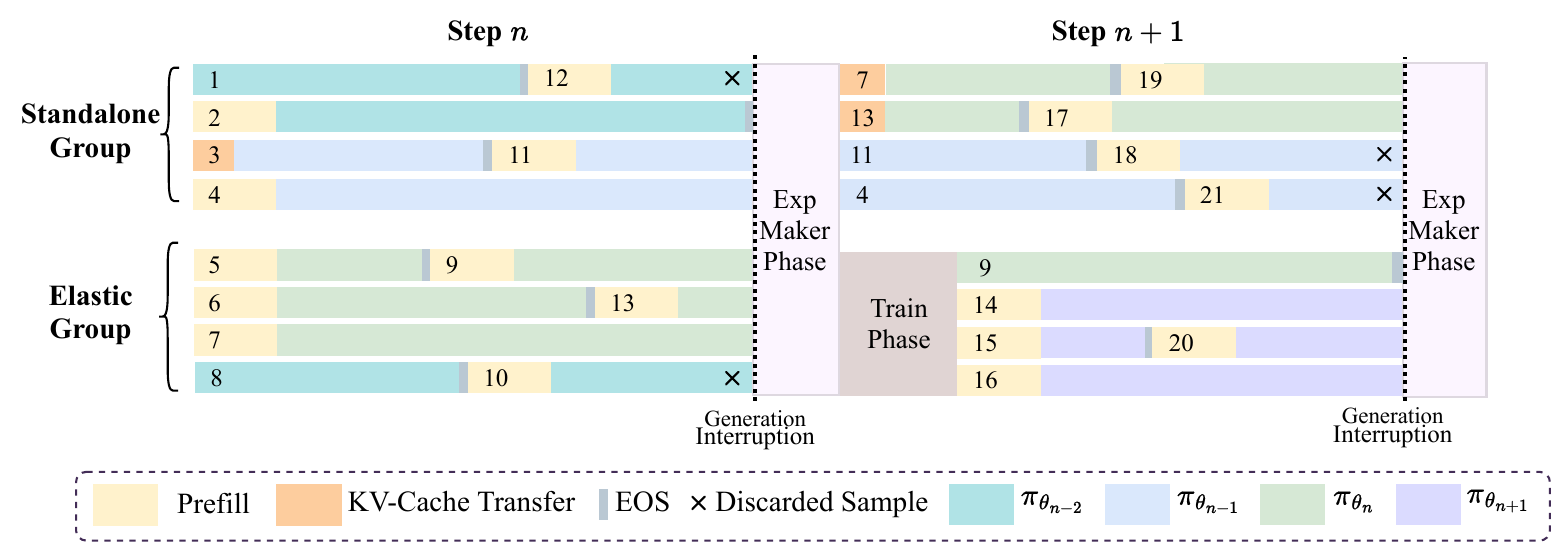}
    \caption{
    The staleness in the demo timeline of DORA system is set to 2, which allows up to three policy weights $\theta_{j}$. At the $n$-th step, prompts 5, 8, 3, 1, 6, and 2 are completed sequentially. The completion of prompt 2 fills the batch, which is then used for training. Prompts 10 and 12 are discarded and scheduled for regeneration. The remaining prompts continue rollout using KV-cache reuse or transfer.}
    \label{fig:streaming_workflow}
\end{figure}

\subsection{RL Infrastructure}
\label{subsec:rl_infra}
The efficiency in RL training is hindered by two primary problems: \textit{RL scheduling}~\citep{xiao2023adaptive, hu2024openrlhf,sheng2025hybridflow}, and \textit{skewed generation problem}~\citep{wu2025llamarl,zhong2025streamrl,seed2025seed1}.
In scheduling, the disaggregated architecture leads to device idleness due to dependencies among different stages.
Conversely, the colocated architecture avoids this by allowing all roles share the same devices, but this efficiency comes at a cost. The tight coupling of hardware for heterogeneous workloads, where generation is memory-bound and training is compute-bound, can lead to suboptimal performance.
The second problem, skewed generation, arises in synchronous training, where the entire batch is blocked by the single longest output. This issue is exacerbated in long-context scenarios, such as reasoning or agentic tool use. Asynchronous training approaches~\citep{team2025kimi,fu2025areal}, like partial rollout, have been proposed to optimize the long-tail generation problem. It breaks up the long responses into segments and utilizes the latest actor model to generate each segment across different iterations. However, we observed considerable inefficiency in the re-prefilling of interrupted samples in practice. The use of latest updated actor models requires re-prefilling of all interrupted samples after concatenating them with previously incomplete responses in the rollout. Furthermore, the use of inconsistent policy versions for different segments of a single response could theoretically harm model convergence.

\subsubsection{DORA: Dynamic ORchestration for Asynchronous rollout} 
To address the aforementioned challenges, we introduce our DORA system. The core idea is to optimize long-tail generation by leveraging multiple old versions of the Actor model through streaming rollout while keeping sampling consistency. To further enhance scheduling efficiency and parallelize both generation and training stages without device idleness, we introduce the elastic colocation of RL roles. As illustrated in Figure~\ref{fig:streaming_workflow}, DORA employs a disaggregated architecture that divides accelerator clusters into two distinct groups:

\begin{itemize}
\item\textbf{Standalone Generator Group}: A pool of devices exclusively dedicated to the Generator role, ensuring optimized rollout. The Generator is a replica of the Actor model specialized for inference.
\item\textbf{Elastic Role Group}: A pool of devices where roles are elastically colocated to ensure flexibility and efficiency. These devices can dynamically switch between serving as Generators and executing various training-related roles (e.g., Reference \& Actor, Reward \& Critic) with elasticity.
\end{itemize}

\textcolor{black}{Based on our resource scheduling for asynchronous rollout, we present the workflow of our DORA system as follows:}

\begin{figure}
    \centering
    \includegraphics[width=1.0 \linewidth]{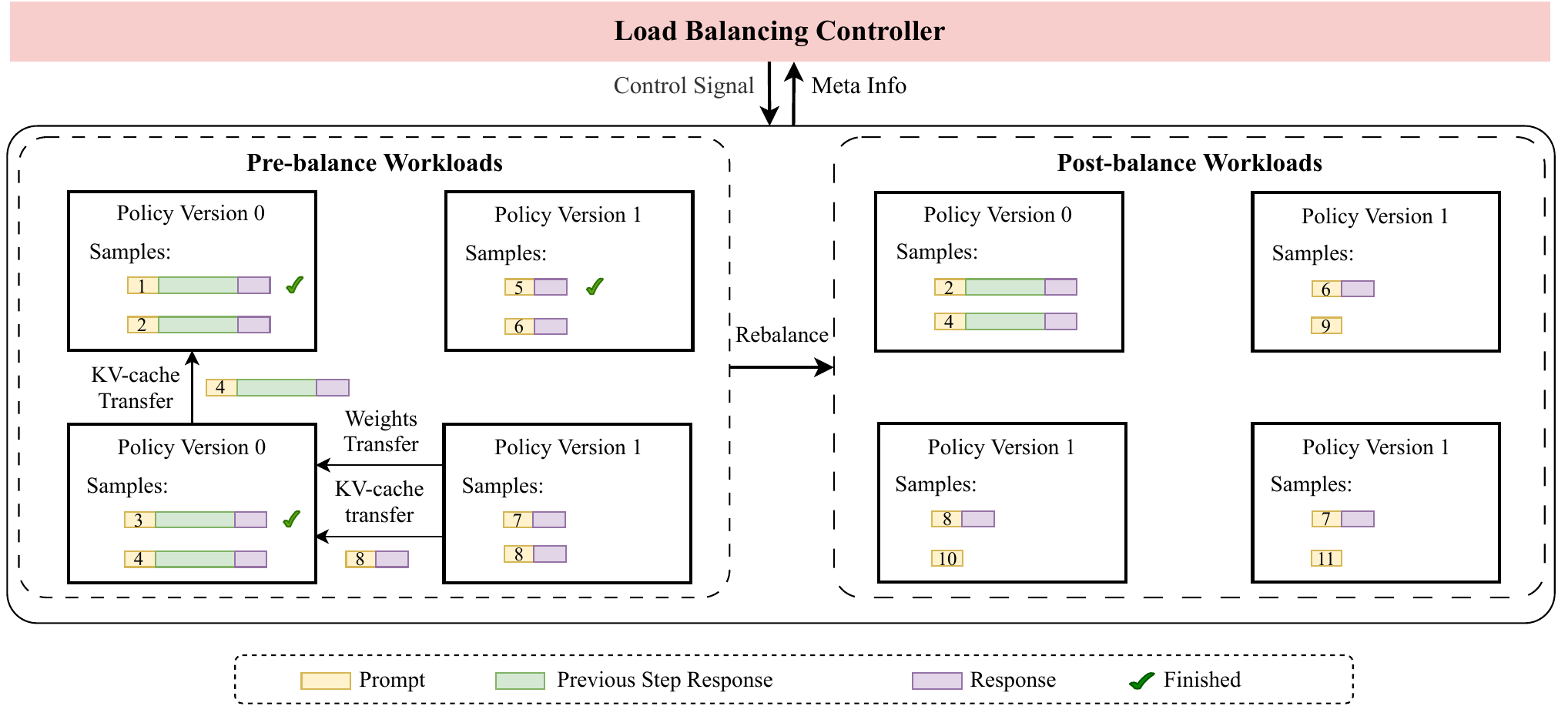}
    \caption{The workflow of Load Balancing. The Load Balancing Controller monitors the load on each device, and when predefined thresholds are met, it initiates resource redistribution, including weight transfer and KV-cache transfer.}
    \label{fig:kv_cache_workflow}
\end{figure}

\noindent\paragraph{Generation Phase} To improve the rollout throughput, Generator devices are scaled up, with both Standalone and Elastic Groups activating the inference engine for rollouts. The inference instances maintain up to a predefined staleness number of policy weight versions. During the rollout stage, our Load Balancing Controller rebalances the resource allocation across various policy versions and reuses the KV-cache within the inference engine, as illustrated in Figure~\ref{fig:kv_cache_workflow}. Crucially, completed samples are immediately streamed to the next stage, without blocking the subsequent stages.
\noindent\paragraph{Experience-Maker Phase} Once the generated samples are satisfied for the training conditions, the Elastic Group scales down its Generator roles, and other RL roles are activated. In a partial colocation setup, Reference \& Actor and the Reward \& Critic roles execute the inference stage in parallel. Meanwhile, the Standalone Generators temporarily switch to the training engine to recompute log probabilities, a critical step to minimize the system-level mismatch between the inference and the training engine. Once completed, the Standalone Group reactivates the Inference engine and continues generation using the previous policy versions. 
\noindent\paragraph{Model Training Phase} Finally, the Actor and Critic models are trained on the collected experience. At the same time, the Standalone Group continues to generate without blocking, while rebalancing workloads and reallocating resources. Notably, the specific policy version will be deleted once it meets the user-defined eviction strategy. After the training finished, the latest policy weights are efficiently synchronized back to the Generator roles using layer-wise Point-to-Point communication, to prepare for the next round of RL training.

The key benefits of DORA are summarized as follows: 
1) The streaming architecture ensures that the responses completed first can be immediately processed in subsequent stages, without being blocked by the longest response. 2) The multi-version design guarantees that each response is generated completely by the same Actor model until completion, eliminating the inconsistency among segments. This also enables easy reuse of the KV-cache for the interrupted samples to significantly reduce the overhead, especially in prefill-heavy scenarios. 
3) Our Elastic Colocation strategy achieves near-zero bubbles in terms of the device idleness except for the negligible duration via intra-process context switching and offloading. It also preserves the advantages of the disaggregated architecture by allowing flexible allocation of both the number and type of accelerators to distinct workloads.

\subsubsection{Optimization for Large-Scale Training} 
To enable industrial-level RL training across tens of thousands of accelerators with our DORA system, we have introduced several key engineering optimizations.

\paragraph{Massive Streaming RPC} The control plane of our system is built on PyTorch RPC ~\citep{damania2023pytorch}, which provides remote procedure calls optimized for the tensor. It decreases the significant serialization and deserialization costs for tensors, and allows dedicated and flexible control over the computing clusters. To enable large-scale RPC capability, we enhanced the TCPStore implementation with additional group-key primitives and data compression during RPC initialization, reducing communication complexity from $O(N^2)$ to $O(N)$. At runtime, we introduced bidirectional streaming RPC—unlike the unary RPC in Pytorch—which enables high-performance streaming rollouts for Inference Engine during asynchronous training.
\paragraph{Efficient MoE Parallelism at Scale}

To deploy LongCat-Flash on our accelerators, we employ a high degree of expert parallelism for generation. 
This strategy not only distributes the computational load but also increases the available memory per accelerator, which is critical for accommodating the large KV-cache required by long-context tasks. 
However, as the scale of expert parallelism increases, maintaining synchronization across distributed accelerators is often bottlenecked by host-side kernel launch overheads, which can lead to execution desynchronization. 
To resolve this, we employ a graph-level compilation approach to reduce the kernel dispatch frequency~\citep{ansel2024pytorch}, thereby enabling the application of graph-level optimizations and effectively overlapping communication with computation. 
As a result, this strategy yielded a $1.5$× rollout speedup, compared to standard eager execution.

Ultimately, the combination of the DORA architecture and large-scale optimizations demonstrates remarkable performance and industrial-grade capability, achieving more than a threefold speedup compared to synchronous training for our 560B LongCat-Flash models across tens of thousands of accelerators.


\subsection{RL Algorithm}

\subsubsection{Training Objective}
Our RL algorithm is developed based on the DORA system. We denote $\pi$ as the autoregressive language model parameterized by $\theta$. Given a query $x$ from the training set $\mathcal{D}$, the likelihood of response $y$ is denoted as $\pi_\theta(y|x)=\prod_{t=1}^{|y|}\pi_\theta(y_t|x,y_{<t})$, where $|y|$ denote the length of the response $y$. Using samples generated from behavior policy $\pi_{\mu}$, Group Relative Policy Optimization (GRPO)~\citep{grpo}, a variant of PPO~\citep{schulman2017proximalpolicyoptimizationalgorithms}, optimizes the policy model within a trust region with group-level advantages, via the following objective:
\begin{equation}
\label{eq:grpo}
\begin{aligned}
\mathcal{J}_\text{GRPO}(\theta) =\ 
&\mathbb{E}_{\substack{x\sim\mathcal{D},\\\{y_i\}_{i=1}^G\sim\pi_{\mu}(\cdot|x)}}\Bigg[ 
    \frac{1}{G} 
    \sum_{i=1}^G \frac{1}{|y_i|}\sum_{t=1}^{|y_i|} \Bigg(
        \min \Big(
            r_{i, t}(\theta) \hat{A}_{i, t},\operatorname{clip}_\varepsilon\big(
                r_{i, t}(\theta)
            \big) \hat{A}_{i, t}
    \Big)-\beta\mathbb{D}_{\text{KL}}[\pi_\theta||\pi_{\text{ref}}]\Bigg) 
\Bigg],
\end{aligned}
\end{equation}
where $r_{i,t}(\theta)=\frac{\pi_{\theta}(y_{i,t}|x,y_{i,<t})}{\pi_{\mu}(y_{i,t}|x,y_{i,<t})}$ is the importance weight, $\operatorname{clip}_\varepsilon$ is the function that clips to $[1-\varepsilon,1+\varepsilon]$ and $\epsilon$ defines the clipping range, 
$\hat{A}_{i,t}$ is the estimated advantage, $G$ represents the sample group from the same query, $\pi_{\text{ref}}$ is the SFT model. However, when this objective is applied to asynchronous training in complex reasoning scenarios, it faces significant challenges due to \textit{distribution drift}, which can disrupt the model's convergence and result in its rapid collapse. This phenomenon can be categorized into two distinct sources:
\begin{itemize}
    \item \textbf{Engine Numerical Gap:} To achieve high throughput and data efficiency, highly optimized inference engines, e.g., vLLM~\citep{vllm}, are naturally applied to generate samples. However, these engines utilize optimizations like kernel fusion that do not guarantee bitwise consistency. This inconsistency is especially critical when the inference and training backends (e.g., the Megatron engine~\citep{shoeybi2019megatron}) are mismatched. While it is possible to use sampling probabilities from the inference engine as $\pi_{\mu}$ during policy optimization, the numerical errors accumulated from this backend mismatch can lead to instability.
    \item \textbf{Policy Staleness:} In asynchronous training, each generated sample may originate from multiple prior versions of the policy, which can become outdated as the current policy $\pi_\theta$ undergoes continuous updates. This discrepancy between the behavior policy that generates the data and the target policy being optimized introduces instability into the training process, hindering convergence and potentially causing model collapse in extreme cases. The standard objective like Eq.~\ref{eq:grpo}, which assumes a high degree of policy alignment, is not robust to these deviations, and the effects of staleness impair its effectiveness.
\end{itemize}
To mitigate the aforementioned issues, we revise the GRPO objective with the following improvements:
\begin{itemize}
    \item The vanilla GRPO loss includes a KL divergence loss term to prevent the policy from deviating far from the reference model. However, with the use of default $k_3$ estimator, the corresponding gradient of this term is biased during optimization despite its unbiased expectation~\citep{zang2025KLGRPO}. Hence, we remove the KL loss term in the GRPO loss, which helps with the substantial policy updates. 
    \item We employ token-level loss, as opposed to sample-level loss, to enhance both the stability of training and the final performance of the model. Furthermore, following the approach of~\citet{dr_grpo}, we utilize the global constant maximum generation length during training as the denominator of the loss function. This adjustment mitigates the length bias that can pose challenges to training robustness.
    \item Setting the clipping range is essential for effective policy optimization, as it affects both exploration and model stability. Besides, as expert routing strategy may change across different versions of policies, the staleness issue can be even more obvious in sparse MoE models, where negative token-level advantages can therefore lead to excessively large importance sampling ratios and unbounded variance. 
    Following~\citet{dapo} and \citet{tencent_moba}, we employ a triplet clipping scheme: $\epsilon_{\text{neg}_\text{low}}$ and $\epsilon_{\text{neg}_\text{high}}$ bound the importance ratio for negative advantages, while $\epsilon_{\text{pos}_\text{high}}$ provides an upper bound for positive advantages. This strategy prevents model collapse and maintains sufficient entropy for effective exploration.
    \item The engine numerical gap can accumulate during RL training and therefore destabilize the whole training procedure. We thus apply truncated importance sampling~\citep{off-policy-rl, ionides2008truncated} to mitigate the distribution mismatch between the inference engine and the train engine.
\end{itemize}
The final objective can be formulated as follows:
\begin{equation}
\begin{aligned}
\mathcal{J}(\theta) =\ 
&\mathbb{E}_{\substack{x\sim\mathcal{D},\\\{y_i\}_{i=1}^G\sim\pi_{\mu}(\cdot|x)}}\Bigg[ 
    \frac{1}{G\cdot T_\text{max}} 
    \sum_{i=1}^G \sum_{t=1}^{|y_i|} 
    \min\bigg(r_{i,t}(\mu),C\bigg)\cdot \\
&\qquad\max\Bigg(
        \min \Big(
            r_{i, t}(\theta) \hat{A}_{i, t},\ 
            \operatorname{clip}\big(
                r_{i, t}(\theta),\ 1-\varepsilon_{\text{neg}_\text{low}},\ 1+\varepsilon_{\text{pos}_\text{high}}
            \big) \hat{A}_{i, t}
        \Big),\ 
        \varepsilon_{\text{neg}_\text{high}}\hat{A}_{i, t}
    \Bigg)
\Bigg],
\end{aligned}
\end{equation}
where $T_{\text{max}}$ is the maximum generation length, $r_{i,t}(\mu)=\frac{\pi_{\mu}^\text{train}}{\pi_{\mu}^\text{infer}}$ is the importance ratio between train and inference engine under the sampling policy $\mu$, and $C$ is a constant value. 

\subsubsection{Efficient Training Strategies}

To better balance effectiveness and efficiency, while maintaining the stability and avoiding reward hacking of the model, we also utilize other techniques:
\noindent\paragraph{With-replacement Online Filtering} We employ online filtering to remove prompts with accuracy scores equal to 1 (fully correct) or 0 (fully incorrect) in the streaming generation stage, retaining the samples with consistently challenging difficulties that provide an effective gradient signal to prevent large gradient fluctuations. To ensure the data consumed-at-least-once and the integrity~\citep{xiao2024antdt}, we develop a sampling-with-replacement strategy for training, differing from the sampling-without-replacement used in dynamic sampling~\citep{dapo}. 
This mechanism enables over-sampling prompts to be regenerated at each training step in synchronous training scenarios. In asynchronous training scenarios, the prompts are reused if their staleness does not exceed the maximum staleness threshold; otherwise, they are regenerated.
\noindent\paragraph{Staleness Control} In the streaming pipeline, we apply maximum staleness as a part of interruption strategy to maintain controllable freshness of the generated samples. To enhance sample efficiency, we apply a data reuse strategy, where the oversampled data of the online filter are stored in the replay buffer and re-sampled in subsequent training iterations, determined by a predefined reuse-ratio. This mechanism caches these stale samples in the replay buffer, allowing them to be proportionally mixed with new samples in subsequent training iterations. Meanwhile, this mixed training batch is required to be shuffled to stabilize the staleness across the training within the buffer. Although this strategy inevitably increases the average staleness, the gains in sample efficiency justify it as an effective trade-off. 
\noindent\paragraph{Incomplete-Signal Masking} We apply a masking strategy to samples that have grading issues, such as sandbox execution errors during code evaluation. This ensures the reliability of the reward signal, resulting in a marginally biased but low-variance gradient. We also mask samples that reach the generation token length without being identified as repetitions. This helps prevent the loss from being affected by outputs that are truncated due to length limits, further improving the stability of the training signal.

\subsection{Reward System}
The reward system is crucial for providing the direction of optimization during the training process. To train LongCat-Flash-Thinking, we develop a well-designed reward system with different reward models to provide accurate reward signals for different tasks.


\noindent\paragraph{Non-Verifiable Tasks}
A discriminative reward model is employed to provide reward signals for non-verifiable tasks such as creative writing, knowledge QA, etc. To obtain this reward model, we initialize it based on the LongCat-Flash SFT checkpoint and subsequently train it on a comprehensive preference dataset, which has been curated through joint annotation by both humans and models. This approach enables the discriminative reward model to accurately capture preferences between different responses. For long CoT responses, we do not include the reasoning process as input; therefore, the reward model evaluates only the answer portion.

\noindent\paragraph{Verifiable Tasks}

For STEM domains, instead of using a rule-based reward system, a Generative Reward Model (GenRM) with reasoning process has been developed to provide the reward signal during the training process~\citep{zhou2025libra}. Given the question, the GenRM compares the reference answer with the response from the LLM and determines whether the response is correct.


There exist several advantages of using the GenRM with reasoning process. Firstly, GenRM accommodates various expressions of answers that have the same meaning, e.g., $a^2 - b^2$ and $(a+b)(a-b)$. At the same time, GenRM is capable of handling complex expressions. Moreover, our GenRM with reasoning process not only provides predictions, but also reveals the reason behind the predictions. The reasoning process enables us to continuously improve the GenRM. We compare the effectiveness of distinct reward models: a rule-based reward method, a non-reasoning GenRM that directly outputs True or False, and our GenRM that incorporates a reasoning process, on a human-labeled test set. Table \ref{tab:reward_models_accuracy} presents the prediction accuracy of these models, demonstrating the effectiveness of our GenRM method.
\begin{table*}[h]
\small
\centering
\caption{Prediction accuracy of different reward models.}
  \begin{tabular}{ccc}
    \toprule
    Rule-based Reward Model & Non-Reasoning GenRM & Reasoning GenRM (Ours) \\
    \midrule
    80.9\%            & 94.0\%              & 98.8\% \\
    \bottomrule
\end{tabular}
\label{tab:reward_models_accuracy}
\end{table*}

For coding tasks, a distributed code sandbox cluster is developed to efficiently manage millions of concurrent code executions for over $20$ programming languages. 
To handle variable workloads from asynchronous RL, we design an asynchronous interface that processes large batches of code, significantly improving throughput by eliminating constant polling. Furthermore, we also optimize efficiency with once-compilation for multi-run executions to reduce overhead, and ensure fast and reliable data transmission and storage with compression and cache sharding.

\subsection{Training Recipe}
Our RL training recipe follows a structured, three-stage methodology designed to cultivate advanced reasoning, comprising: 1) Domain-Parallel Training, where expert models are independently trained on curated datasets for distinct domains (e.g., STEM, Code, Agentic); 2) Model Fusion, a novel technique to integrate these experts into a single, cohesive agent and consolidate their skills; and 3) General RL Fine-Tuning, a final stage to harmonize the model's capabilities and ensure robust performance across diverse applications.

\subsubsection{Reasoning-oriented RL: A Domain-Parallel Approach}
\begin{figure}[t]
\centerline{\includegraphics[width=0.8\linewidth]{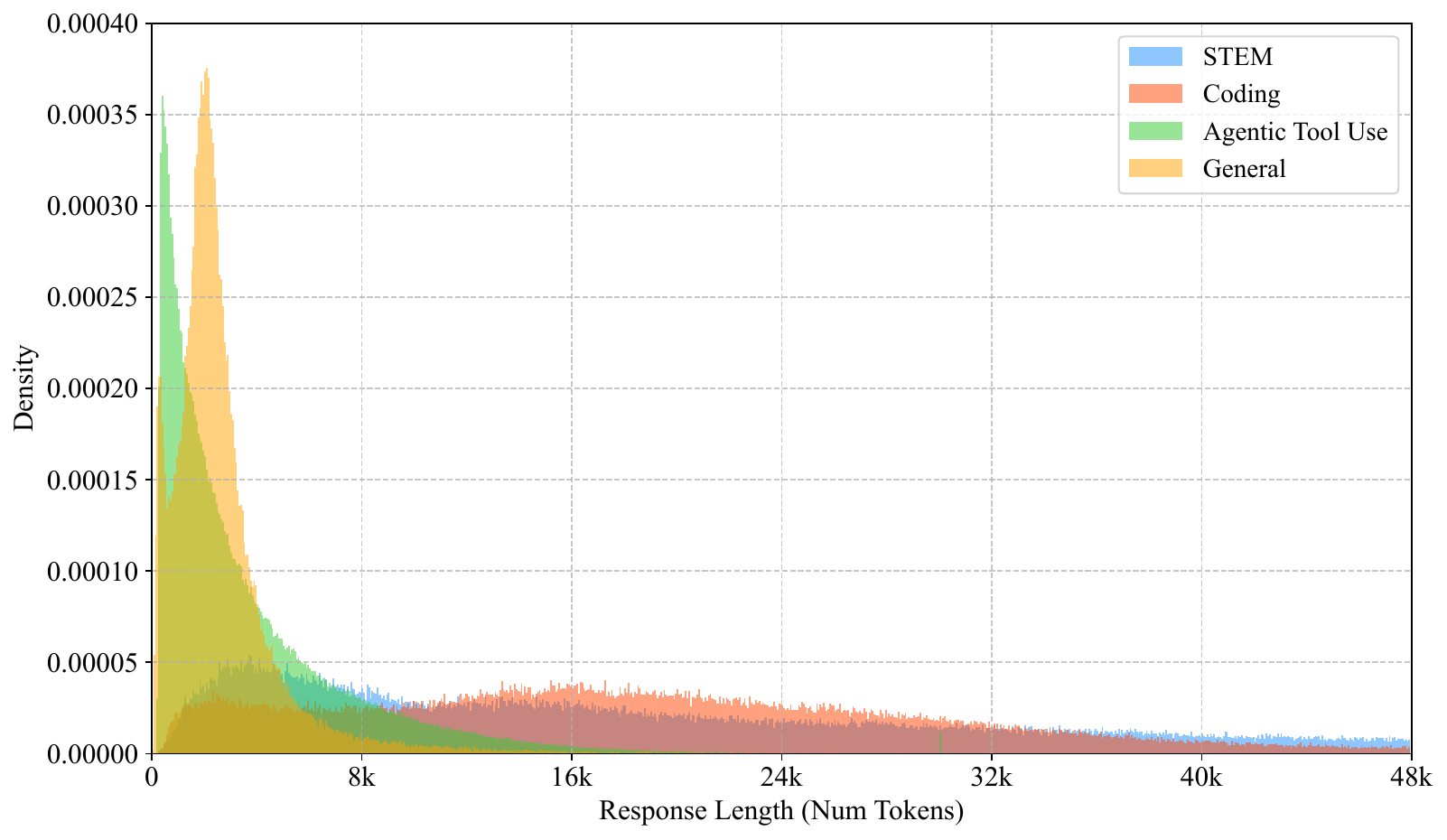}}
\caption{\small
The response length distributions of different domains during large-scale RL training.
}
\label{fig:domain_train_length_dis}
\end{figure}

In large-scale RL, we observed that a domain-mixed training pipeline often results in negative transfer during asynchronous training, leading to inefficiency and sub-optimal performance. 
We attribute this to significant distributional shifts between training batches, caused by varying response characteristics across domains (as shown in Figure~\ref{fig:domain_train_length_dis}).
While sequential training (i.e., optimizing one domain at a time) can partially mitigate this issue, it is inherently inefficient and inflexible.
Once later training stages commence, it is difficult to revisit or refine capabilities from earlier domains. Therefore, we introduce a domain-parallel training framework. This approach first trains separate "expert" models for distinct reasoning domains and then merges them into a single, powerful model that achieves nearly Pareto-optimal performance across all specializations. 
The process concludes with a general RL phase to ensure broad capabilities and alignment. 
This overall pipeline is illustrated in Figure~\ref{fig:parallel_training_pipeline}.

\noindent\paragraph{Query Curation for RL}
To supply high-quality data for the RL stage, we implemented a rigorous, multi-faceted curation protocol tailored for each reasoning domain.
For STEM and Code queries, the protocol begins with standard decontamination and deduplication against known benchmarks. 
We further refine the STEM dataset by excluding unsuitable formats, such as multi-part, multiple-choice, or true/false questions. 
For Code queries, test cases are systematically reformatted into a standardized input-output structure to ensure compatibility. 
A crucial subsequent filtering step is applied to both domains to prevent reward signal bias: using an SFT model, we generate multiple responses for each query and retain only those instances exhibiting a balanced distribution of correct and incorrect solutions. 
This avoids problems that are trivially easy (all correct) or impossibly hard (all incorrect), thereby improving training efficacy. Specifically for Code, we also leverage sandbox execution feedback to identify and remove ambiguous problems or mismatched test cases that could lead to false negatives.
For Agentic RL, we curated a specialized dataset focused on mathematical problems that necessitate complex reasoning and tool application. 
Each training instance is structured as a triplet, containing the problem description, a reference answer, and an accompanying grading rubric. 
This detailed structure is designed to effectively guide the model in learning the appropriate tool-use trajectories for solving intricate tasks.

\noindent\paragraph{Domain-Parallel Training}
A key advantage of our domain-parallel approach is the ability to tailor training methodologies to the unique characteristics of each reasoning domain. 
We applied distinct configurations for STEM, Code, and Agentic RL to maximize their respective strengths:


\begin{itemize}
    \item\textbf{STEM RL:} The training process utilizes a fixed $64$K context length. We implement a curriculum learning strategy by gradually increasing data difficulty (by lowering the pass-rate threshold for inclusion). Concurrently, we dynamically adjust the PPO clipping bound $\varepsilon_{\text{pos}_{\text{high}}}$ to maintain training stability. These approaches guarantee that the model's learning evolves in an efficient manner, while seamlessly adapting to the growing complexity of the training data.
    \item\textbf{Code RL:} We employ a multi-stage curriculum for context length, starting at $48$K tokens, then progressively extending to $56$K, and ultimately reaching to $64$K. The context window is expanded once the $90$th percentile of generated output lengths approaches the current limit, ensuring smooth adaptation.
    \item\textbf{Agentic RL:} The training process utilizes a fixed $48$K context length. We enforce structured reasoning and tool use via two techniques: 1) structured dialogue templates using  \texttt{<longcat\_think>} and \texttt{<longcat\_tool\_call>} tags, and 2) a tool-call format reward that incentivizes syntactically correct tool usage, ensusing stable and interpretable multi-turn trajectories.
\end{itemize}

\subsubsection{Model Fusion}
\begin{figure}[t]
\centerline{\includegraphics[width=0.9\linewidth]{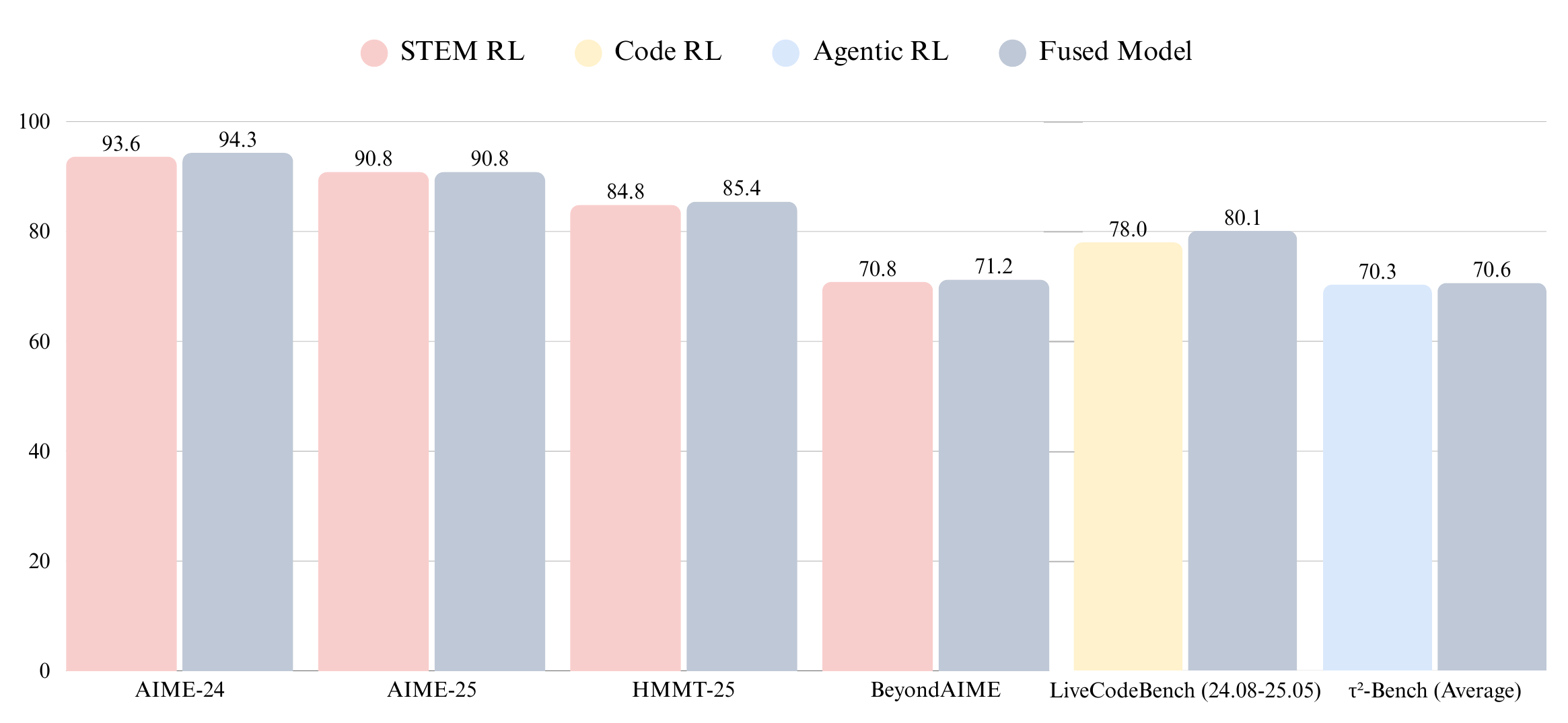}}
\caption{\small
The performance (\%) of the fused model after STEM RL, Code RL, and Agentic RL.
}
\label{fig:model_fusion_results}
\end{figure}

To consolidate the capabilities of our domain-expert models, we merge their parameters into a single, unified agent. 
This approach is supported by prior work~\citep{yadav2023ties, DBLP:conf/naacl/LeeLWWCW25}, which has shown that merging domain-specific models can yield a single model with superior overall performance. 
The primary challenge is mitigating parameter interference between the experts. 
To address this, we employ a three-pronged strategy inspired by recent advances:
1)~\textbf{Normalization}: We normalize the magnitude of task vectors ($\tau_{i}=\theta^{i}_{RL}-\theta_{SFT}$) to balance contributions from different domains.
2)~\textbf{Dropout}: Similar to DARE~\citep{yu2024language}, we apply dropout to prune redundant delta parameters.
3)~\textbf{Erase}: Inspired by SCE~\citep{DBLP:journals/corr/abs-2408-07990}, we erase parameter elements with minority-direction updates.
This fusion strategy constructs a single model that excels in mathematical reasoning, coding, and agentic capabilities, as demonstrated in Figure~\ref{fig:model_fusion_results}.

\subsubsection{Final Aligenment with General RL}
The final stage of our pipeline is a general RL phase designed to enhance the model's capabilities in broad scenarios (e.g., creative writing, instruction following) and prevent any regression of core competencies like safety after fusion. 
We first compile a diverse dataset from open-source and synthetic queries, then apply clustering algorithms to deduplicate and filter for high-quality, challenging data. 
This curated dataset is then used for a final round of PPO training, ensuring the model is well-aligned, robust, and adaptable for real-world applications.



\section{Evaluations}
In this section, we evaluate our~{\longcatthink} model after the whole training pipeline on automatic benchmarks.
These benchmarks are categorized into several dimensions, including general QA, alignment, mathematics, general reasoning, coding, agentic tool use, formal theorem proving, and safety.

\begin{table*}[t]
\setlength{\tabcolsep}{4pt}
\centering
\caption{Performance (\%) comparison across multiple benchmarks (Best in \textbf{bold}, second best is \underline{underlined}).$\dagger$ indicates the score is from external reports.}
\label{tab:main_results}
\resizebox{0.98\textwidth}{!}{
\begin{tabular}{@{}l| c c c c c c c@{}}
\toprule
\multirow{3}{*}{\textbf{Benchmark}} & \multicolumn{3}{c}{\textit{Open-Weights Reasoning Models}} & \multicolumn{3}{c}{\textit{Close-Weights Reasoning Models}} & \textit{Ours} \\
\cmidrule(lr){2-4} \cmidrule(lr){5-7} \cmidrule(lr){8-8}
& DeepSeek-V3.1- & Qwen3-235B-A22B- & GLM-4.5 & OpenAI-o3 & Gemini2.5-Pro & GPT-5- & \textcolor{mygreen}{LongCat-Flash-} \\
&Thinking &Thinking-2507 & & & &Thinking &\textcolor{mygreen}{Thinking}  \\
\midrule
Architecture &MoE &MoE &MoE&-&-&-&MoE \\
\# Total Params &671B & 235B &355B &- &- &- &560B\\
\# Activated Params &37B &22B &32B &- &- &- &27B \\
\midrule
\rowcolor{mygray}

\multicolumn{8}{c}{\textit{General QA}} \\
MMLU-Pro$_\text{(acc)}$ &84.4 &84.4 &81.5 &\underline{85.3} &\textbf{86.7} &84.5 & 82.6\\
MMLU-Redux$_\text{(acc)}$ &90.5 &91.4 &89.9 &\textbf{93.1} &90.1 &\underline{92.6} & 89.3\\
\midrule
\rowcolor{mygray}

\multicolumn{8}{c}{\textit{Alignment}} \\
IFEval$_\text{(strict prompt)}$ &86.3 &89.3 &85.4 &90.2 &\underline{92.4} &\textbf{92.8} & 86.9\\

Arena-Hard$_\text{(hard prompt gemini)}$ &57.1 &74.5 &67.7 &\underline{87.1}&\underline{87.1}&\textbf{87.7} &69.9\\
\midrule
\rowcolor{mygray}

\multicolumn{8}{c}{\textit{Mathematical Reasoning}} \\
MATH-500$_\text{(Mean@1)}$ &98.8 &\textbf{99.6} &95.4 &98.4 &98.0 &\underline{99.2} &\underline{99.2}\\
HMMT-25$_\text{(Mean@32)}$ &80.4 &\underline{83.8} &76.3 &71.9 &79.3 &\textbf{84.8} &83.7\\
AIME-24$_\text{(Mean@32)}$ &\textbf{93.9} &\textbf{93.9} &89.3 &91.6$^\dagger$ &90.7 &92.0 &\underline{93.3}\\
AIME-25$_\text{(Mean@32)}$ &87.9 &\underline{92.5} &85.5 &88.9$^\dagger$ &89.2 &\textbf{94.6}$^\dagger$ &90.6\\
BeyondAIME$_\text{(Mean@10)}$ &\textbf{71.8} &\underline{71.5} &66.0 &63.2 &63.0 &70.0 &69.5\\
\midrule
\rowcolor{mygray}

\multicolumn{8}{c}{\textit{General Reasoning}} \\
GPQA-Diamond$_\text{(Mean@16)}$ &\underline{84.2} &80.4 &78.3 &81.9 &84.0 &\textbf{84.4} & 81.5\\
ZebraLogic$_\text{(Mean@1)}$ &\underline{96.1}&\textbf{97.5} &90.9 &94.3 &92.4 &92.7 &{95.5}\\
Sudoku-Bench$_\text{(Mean@1)}$ &1.0 &2.0 &1.0 &\textbf{70.0} &0.0 &\underline{63.0} &56.0\\
ARC-AGI$_\text{(Mean@1)}$ &37.5 &45.3 &21.4 &{47.3} &46.8 &\textbf{59.0} & \underline{50.3}\\
\midrule
\rowcolor{mygray}

\multicolumn{8}{c}{\textit{Coding}} \\
LCB (24.08-25.05)$_\text{(Mean@4)}$ &73.5 &75.4 &61.1 &76.2 &74.2 &\textbf{80.6} &\underline{79.4}\\
OJBench$_\text{(Mean@1)}$ &33.6 &32.1 &19.0 &38.4 &\textbf{41.6} &34.1 &\underline{40.7}\\
\midrule
\rowcolor{mygray}

\multicolumn{8}{c}{\textit{Agentic Tool Using}} \\
SWE-Bench$_\text{(Pass@1)}$ &66.0$^\dagger$ &34.4 &64.2$^\dagger$ &\underline{69.1}$^\dagger$ &59.6$^\dagger$ &\textbf{74.9}$^\dagger$ &59.4\\

BFCL V3$_\text{(full)}$ &55.4 &\underline{75.7} &\textbf{79.1} &72.4$^\dagger$ &63.2 &60.1 &74.4\\

$\tau^2$-Bench-Retail$_\text{(Mean@4)}$ &65.4 &68.2 &69.3 &\underline{72.8} &70.9 &\textbf{81.1}$^\dagger$ &{71.5}\\
$\tau^2$-Bench-Airline$_\text{(Mean@4)}$ &44.0 &58.0 &\underline{66.0} &62.5 &58.0 &{62.6}$^\dagger$ &\textbf{67.5}\\
$\tau^2$-Bench-Telecom$_\text{(Mean@4)}$ &23.7 &47.3 &56.1 &67.5 &38.3 &\textbf{96.7}$^\dagger$ &\underline{83.1}\\

VitaBench &13.5 &21.5 &26.8 &\textbf{35.3} &24.3 &29.3 & \underline{29.5} \\
\midrule
\rowcolor{mygray}

\multicolumn{8}{c}{\textit{Formal Theorem Proving}} \\
MiniF2F-Test$_\text{(Pass@1)}$ &\underline{49.6} &11.9 &10.9 &15.2 &13.9 &21.4 &\textbf{67.6}\\
MiniF2F-Test$_\text{(Pass@8)}$ &\underline{74.4} &20.9 &22.1 &29.6 &29.4 &39.7 &\textbf{79.4}\\
MiniF2F-Test$_\text{(Pass@32)}$ &\underline{79.5} &26.6 &27.0 &37.7 &41.8 &51.2 &\textbf{81.6}\\
\midrule
\rowcolor{mygray}

\multicolumn{8}{c}{\textit{Safety}} \\
Harmful &79.2 &\underline{84.3} &70.4 &64.8 &44.3 &56.8 &\textbf{93.7}\\
Criminal &89.7 &\underline{92.7} &88.8 &85.7 &77.4 &87.3 &\textbf{97.1}\\
Misinformation &\underline{81.1} &80.9 &67.1 &42.7 &31.0 &41.9 &\textbf{93.0}\\
Privacy &96.2 &\textbf{100.0} &97.6 &\textbf{100.0} &95.0 &\underline{98.8} &\underline{98.8}\\
\bottomrule
\end{tabular}
}
\end{table*}

\subsection{Benchmarks and Configurations}
The evaluation employs the following benchmarks:
\begin{itemize}
    \item\textbf{General QA: }MMLU-Pro \citep{wang2024mmluprorobustchallengingmultitask}, a robustly re-evaluated version of MMLU \citep{hendrycks2021measuringmassivemultitasklanguage} that corrects errors and reduces contamination. MMLU-Redux \citep{gema2025mmluredux}, another high-quality variant of the MMLU Benchmark. 
    \item\textbf{Alignment: }IFEval \citep{zhou2023ifeval}, an instruction following benchmark consisted of a set of prompts with programmatically verifiable constraints, offering an objective score on the model's fidelity to complex instructions. Arena-Hard \citep{arenahard2024}, a benchmark sourced from the Chatbot Arena platform for assessing model's helpfulness and conversational quality on difficult, open-ended user queries. 
    \item\textbf{Mathematical Reasoning: }Olympiad-level mathematical benchmarks, including MATH-500 \citep{math500}, HMMT-25\citep{HMMT25} (Harvard-MIT Mathematics Tournament), AIME-24 \citep{AIME24} and AIME-25 \citep{AIME25} (American Invitational Mathematics Examinations), and BeyondAIME \citep{bytedanceseed2025beyondaime}.
    \item\textbf{General Reasoning: }GPQA-Diamond \citep{rein2024gpqa}, a benchmark evaluating deep reasoning on graduate-level questions across several science domains. ZebraLogic \citep{lin2025zebralogic}, composed of classic logic grid puzzles that require multi-step deductive reasoning and constraint satisfaction. Sudoku-Bench \citep{seely2025sudoku-bench}, symbolic and structured reasoning by requiring models to solve Sudoku puzzles\footnote{We evaluate the model on the challenging nikoli\_100 dataset.}. ARC-AGI \citep{chollet2019github}, a benchmark designed to measure fluid intelligence.
    \item\textbf{Coding: }LiveCodeBench (LCB) \citep{jain2025livecodebench}, a dynamic benchmark and we evaluate on problems between 2408 and 2505. 
    OJBench~\citep{wang2025ojbenchcompetitionlevelcode}, ACM-ICPC level code reasoning benchmarks. 
    \item\textbf{Agentic Tool Using: } 
    SWE-Bench~\citep{jimenez2024swebench}, sourced from real GitHub issues for evaluating a model's ability to solve software engineering problems. 
    BFCL~\citep{BFCL}, 
    and $\tau^2$-Bench~\citep{barres2025tau2} are Tool-Augmented Reasoning benchmarks. VitaBench~\citep{meituan2025longcat_flash_chat}, our self-designed comprehensive real-world benchmark for systematically evaluating models’ capabilities in addressing complex real-world tasks. 
    \item\textbf{Formal Theorem Proving: }MiniF2F~\citep{zheng2021minif2f} is a benchmark for formal theore proving, featuring a collection of problem statements translated across multiple formal systems. The problems are curated from high school and undergraduate mathematics exercises to challenging questions from prestigious competitions such as the AMC, AIME, and IMO. We evaluate on the test set that consists of 244 problems.
    \item\textbf{Safety}: Following LongCat-Flash-Chat~\citep{meituan2025longcat_flash_chat}, we evaluate the safety performance of LongCat-Flash-Thinking on four major risk categories: \textbf{Harmful} (e.g., violence, hate Speech, insulting, harassment, self-harm, and adult content), \textbf{Criminal} (e.g., illegal activities, terrorism, and underage violations), \textbf{Misinformation} (e.g., misinformation， disinformation, and unsafe practices), and \textbf{Privacy} (e.g., privacy violation and infringement). For each category, we curated a substantial set of private test queries, which then underwent a rigorous manual review to validate their categorization and ensure their quality.

\end{itemize}

For each benchmark category, we employ the following specialized metrics and configurations:
\begin{itemize}
    \item\textbf{General QA: }We use accuracy as the primary evaluation metric. Following~\citep{meituan2025longcat_flash_chat}, we employ a scoring model to assess the semantic alignment between model responses and reference answers. Because this approach recognizes responses that are semantically correct but not textually identical, our reported accuracy scores may be slightly higher than those originally documented.
    
    \item\textbf{Alignment: }To assess instruction compliance, we design regular expressions tailored to the specific rules of each task. This verification process is further supported by both rule-based and model-based answer span extraction tools.
    \item\textbf{Mathematical Reasoning: } For the AIME and HMMT benchmarks, we report the average accuracy over 32 samples (Mean@32), while for MATH-500 
    and BeyondAIME we report Mean@1 and Mean@10, respectively.
    
    \item\textbf{General Reasoning: }We apply the LLM-as-a-Judge for GPQA-diamond, while apply rule-based matching for ZebraLogic. For the GPQA-diamond benchmark, we report the average accuracy over 16 samples (Mean@16), while for ZebraLogic, Sudoku-Bench 
    and ARC-AGI we report Mean@1 accuracy.
    \item\textbf{Coding:} We use the acceptance (AC) rate to evaluate coding performance, where the model scores 1 on one problem only if its code passes all test cases, otherwise it scores 0. For LCB, we use Python 3.11 as the paper suggested. 
    For each benchmark, we utilize official benchmark frameworks to ensure fairness and reproducibility. 
    \item\textbf{Agentic Tool Using: } Similar to the task of coding, we also utilize official benchmark frameworks to ensure fairness and reproducibility. 
    \item\textbf{Formal Theorem Proving: } We report the $pass@k$ metric ($k\in\{1, 8, 32\}$) for MiniF2F-Test. The proof will be accepted if it passes the syntax check through the Lean4 server.
    \item\textbf{Safety: }We directly choose accuracy as the primary metric.
\end{itemize}

\subsection{Evaluation Results}
As illustrated in Table~\ref{tab:main_results}, we compare~{\longcatthink} with several advanced reasoning models, including DeepSeek-V3.1-Thinking~\citep{deepseekai2025deepseekv3technicalreport}, Qwen3-235B-A22B-Thinking~\citep{qwen3_thinking_2507}, GLM-4.5~\citep{zeng2025glm}, OpenAI-o3~\citep{o3}, Gemini2.5-Pro~\citep{gemini25pro}, and GPT-5-Thinking~\citep{gpt5}.
The results of our comprehensive evaluation indicate that~{\longcatthink} is a highly capable and versatile model. It consistently demonstrates superior performance across a broad spectrum of reasoning tasks, surpassing its counterparts that require a greater number of activated parameters. A detailed breakdown of these capabilities is provided in the subsequent analysis.
The inference parameters of our LongCat-Flash-Thinking are set as temperature=1.0, topk=-1, and topp=0.95.

\begin{figure}[t]
\centerline{\includegraphics[width=0.85\linewidth]{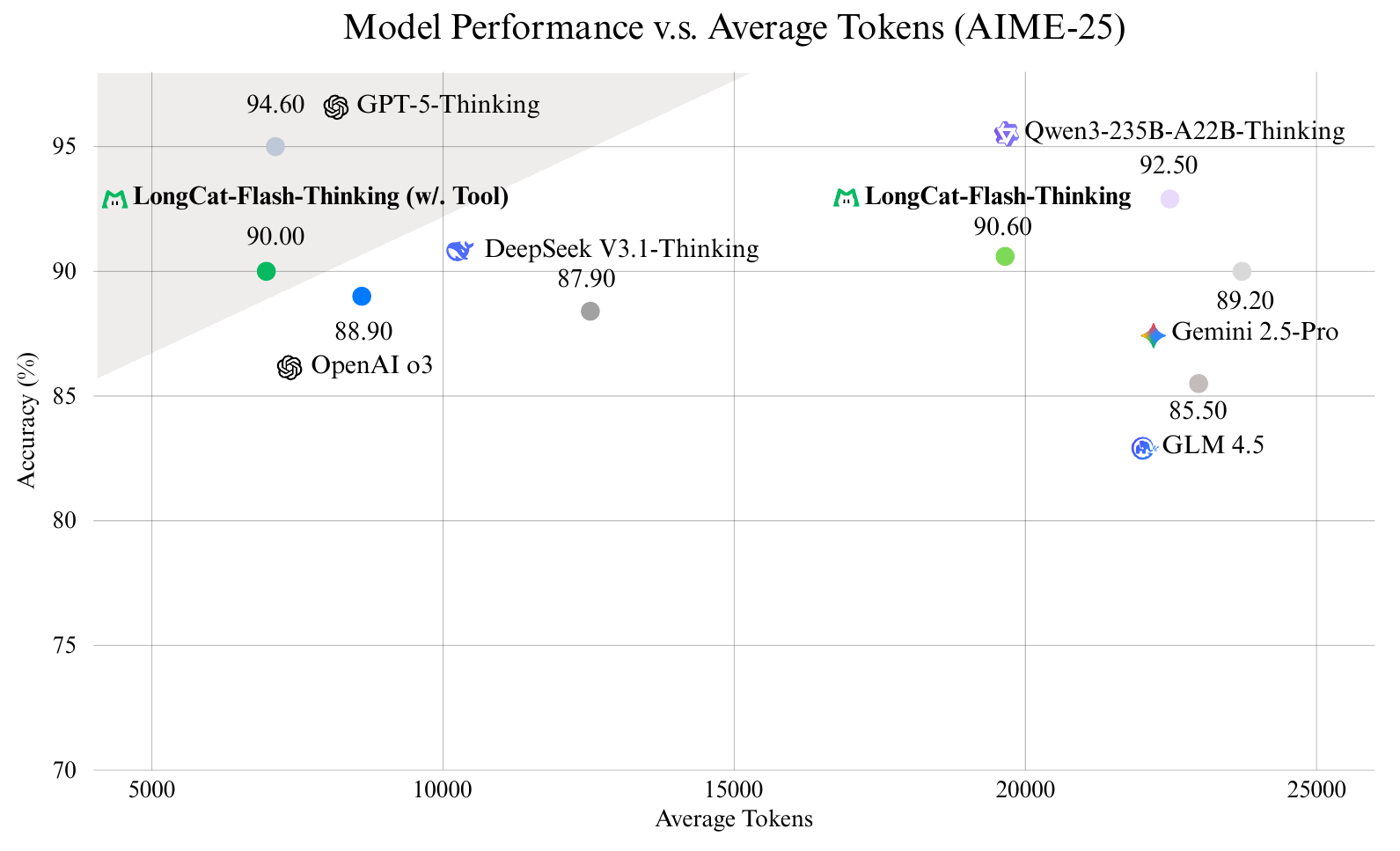}}
\caption{\small
The comparison of performance (\%) and average tokens on AIME-25.
}
\label{fig:token_efficiency}
\end{figure}

\begin{itemize}
    \item\textbf{General Abilities: }In the domain of general knowledge, the result demonstrates robust foundational understanding ability of~{\longcatthink}, particularly on comprehensive multi-task benchmarks. 
    It achieves an accuracy of 89.3\% on MMLU-Redux, rivaling the state-of-the-art open-source model Qwen3-235B-A22B, and maintains strong competitiveness on MMLU-Pro with a score of 82.6\%. 
    These results are especially remarkable considering~{\longcatthink}’s efficiency.
    
    \item\textbf{Alignment Abilities: }In alignment tasks, LongCat-Flash-Thinking exhibits strong and competitive performance. It attains an impressive score of 86.9\% on the IFEval benchmark (strict prompt) and 69.9\% on Arena-Hard(hard prompt gemini), demonstrating robust capability in following complex instructions and surpassing several key baselines, including DeepSeek V3.1.
    
    \item\textbf{Mathematical Abilities: }{\longcatthink} demonstrates outstanding proficiency in mathematical reasoning, distinguishing itself as one of the top-performing models currently available. In particular, it achieves a highly competitive score of 99.2\% on the MATH500 benchmark, and its true capabilities become even more evident on more challenging benchmarks. The model delivers impressive results on HMMT and AIME-related benchmarks, outperforming strong baselines such as OpenAI-o3 and rivaling leading proprietary models like Qwen3-235B-A22B and GPT-5. These findings highlight its advanced ability to solve complex, multi-step problems.
    \item\textbf{General Reasoning Abilities: }{\longcatthink} demonstrates superior general reasoning capabilities, particularly in tasks that require structured logic. It achieves exceptional performance on ARC-AGI, attaining a score of 50.3\% and surpassing leading proprietary models such as OpenAI-o3 and Gemini2.5-Pro. Additionally, it displays remarkable aptitude in puzzle reasoning, with scores of 56.0\% on Sudoku-Bench and 95.5\% on ZebraLogic, highlighting its advanced ability to solve complex, non-linguistic puzzles. 
    
    \item\textbf{Coding Abilities: }In the coding domain,~{\longcatthink} 
    demonstrates state-of-the-art performance and strong overall capabilities. On LiveCodeBench, it achieves a score of 79.4\%, significantly outperforming all listed open-source models and performing competitively with the top proprietary model, GPT-5 (80.6\%). This showcases its excellent proficiency in solving fresh, competitive programming problems. 
    Furthermore, it scores 40.7\% on OJBench, surpassing most proprietary models and nearly matching the leading score of Gemini2.5-Pro (41.6\%).
    
    \item\textbf{Agentic Abilities: }The evaluation of agentic capabilities reveals that~{\longcatthink} excels in complex, tool-augmented reasoning. 
    demonstrates robust capabilities in agentic tool using. Specifically, it achieves a state-of-the-art result on $\tau^2$-Bench-Airline with a score of 67.5\% and delivers highly competitive performance across a range of other benchmarks, including SWE-Bench, BFCL V3, and VitaBench. 
    In particular, enabling targeted tool use substantially improves the trade-off between performance and token budget. 
    In our experiments,~{\longcatthink} with the tool reduces average token consumption from $19,653$ to $6,965$ ($\approx$64.5\% less) while preserving AIME-25 accuracy, as shown in Figure \ref{fig:token_efficiency}. This demonstrates that agentic systems can achieve nearly state-of-the-art performance not only by extending model-only reasoning but also efficiently by offloading select reasoning steps to external tools.
    
    \item\textbf{Formal Reasoning for ATP: } 
    In the domain of proving,~{\longcatthink} demonstrates state-of-the-art capabilities. On the MiniF2F-Test benchmark, it achieves a pass@1 score of 67.6\%, outperforming the second best model, DeepSeek V3.1 by a substantial margin of 18\%. This lead is maintained across pass@8 and pass@32, highlighting its superior ability in generating structured proofs and formal mathematical reasoning.
    
    \item\textbf{Safety Abilities: }On safety benchmarks,~{\longcatthink} demonstrates state-of-the-art performance in refusing to answer harmful or unethical queries. It achieves the overall best performance, significantly outperforming all other evaluated open-source and proprietary models. Its strength is consistent across all sub-categories, with top scores in Harmful (93.7\%), Criminal (97.1\%), and Misinformation (93.0\%). This robust performance underscores the effectiveness of our model's safety alignment and its reliability in adhering to safety protocols.
\end{itemize}





\section{Conclusion}

In this work, we introduce~{\longcatthink}, a $560$-billion-parameter Mixture-of-Experts (MoE) reasoning model, which achieves state-of-the-art performance among open-source models on multiple reasoning tasks with exceptional efficiency. 
The core innovations underpinning {\longcatthink} are as follows:
1) A well-designed cold-start training strategy which significantly enhances the reasoning potential and equips the model with specialized skills in both formal and agentic reasoning.
2) A domain-parallel training scheme that decouples optimization across STEM, coding, and agentic tasks, enabling the merge of the resulting expert models into a nearly pareto-optimal model.
3) An efficient and scalable RL framework built on the proposed Dynamic ORchestration for Asynchronous rollout (DORA) system, thereby achieving industrial-scale asynchronous training on tens of thousands of accelerators.
We hope that the open-sourcing of~{\longcatthink} will advance research in reasoning models, particularly in the areas of high-quality data strategies, efficient RL training, and native agentic reasoning.

\newpage

\section{Contributions}

The listing of authors is in alphabetical order.
Entries with identical English names will be sorted based on the order of Chinese stroke count and pronunciation.
An asterisk (*) indicates members who have departed from the team.

\begin{CJK}{UTF8}{gbsn}

\begin{center}
\begin{tabular}{p{0.185\textwidth}p{0.185\textwidth}p{0.185\textwidth}p{0.185\textwidth}p{0.185\textwidth}}
Anchun Gui & Jiahao Liu & Meng Zhou & Weifeng Tang & Yifan Lu \\
Bei Li & Jiahuan Li & Mengshen Zhu & Wenjie Shi & Yiyang Li \\
Bingyang Tao & Jialin Liu & Peng Pei & Wenlong Zhu & Youshao Xiao \\
Bole Zhou & Jianfei Zhang & Pengcheng Jia & Xi Su & Yuanzhe Lei \\
Borun Chen & Jianhao Xu & Qi Gu & Xiangcheng Liu & Yuchen Xie \\
Chao Zhang (张超) & Jianing Wang & Qi Guo & Xiangyu Xi & Yueqing Sun \\
Chao Zhang (张朝) & Jiaqi Sun & Qiong Huang & Xiangzhou Huang & Yufei Zhang \\
Chengcheng Han & Jiaqi Zhang & Quan Chen & Xiao Liu & Yuhuai Wei \\
Chenhui Yang & Jiarong Shi & Quanchi Weng & Xiaochen Jiang & Yulei Qian \\
Chi Zhang & Jiawei Yang & Rongxiang Weng & Xiaowei Shi & Yunke Zhao \\
Chong Peng & Jingang Wang & Ruichen Shao & Xiaowen Shi & Yuqing Ding \\
Chuyu Zhang & Jinrui Ding & Rumei Li & Xiaoyu Li & Yuwei Jiang \\
Cong Chen & Jun Kuang & Shanglin Lei & Xin Chen & Zhaohua Yang \\
Fengcun Li & Jun Xu* & Shuai Du & Xinyue Zhao & Zhengyu Chen \\
Gang Xu & Ke He & Shuaikang Liu & Xuan Huang & Zhijian Liu \\
Guoyuan Lin & Kefeng Zhang & Shuang Zhou & Xuemiao Zhang & Zhikang Xia \\
Hao Jiang & Keheng Wang & Shuhao Hu & Xuezhi Cao & Zhongda Su \\
Hao Liang & Keqing He* & Siyu Xu & Xunliang Cai & Ziran Li \\
Haomin Fu & Li Wei & Songshan Gong* & Yajie Zhang & Ziwen Wang \\
Haoxiang Ma & Liang Shi & Tao Liang & Yang Chen & Ziyuan Zhuang \\
Hong Liu & Lin Qiu & Tianhao Hu & Yang Liu (刘阳) & Zongyu Wang \\
Hongyan Hao & Lingbin Kong & Wei He & Yang Liu (刘洋) & Zunyuan Yang \\
Hongyin Tang & Lingchuan Liu & Wei Shi & Yang Zheng & LongCat-Flash \\
Hongyu Zang & Linsen Guo & Wei Wang & Yaoming Wang &  \\
Hongzhi Ni & Longfei An & Wei Wu & Yaqi Huo &  \\
Hui Su & Mai Xia & Wei Zhuo & Yerui Sun &  \\
\end{tabular}
\end{center}

\end{CJK}

\newpage

\bibliographystyle{unsrtnat}
\bibliography{references}

\newpage
\appendix
\section{Appendix}

\subsection{Details of Mid-training}
\label{appendix:rbe}

For the mid-training phase,
data quality and diversity are crucial for enhancing the reasoning capabilities of models \citep{xi2025samplemix}. During the filtering phase, we employ a combination of heuristic rules and an LLM-as-a-Judge approach to ensure the solvability and difficulty distribution of queries, as well as the quality and correctness of answers.

To guarantee query solvability, rule-based methods such as URL filtering, multiple tables filtering, and HTML/XML tags filtering are applied to eliminate reasoning problems that rely on external information or are inherently unanswerable. For synthetic answers, we first apply rules to remove those with issues such as repetitive generation, truncation, language mixing, or non-compliance with our desired reasoning format.
To verify answer correctness, we combine model-based and rule-based methods, comparing synthetic responses against gold answers to assess equivalence across various forms and stylistic variations. 

For the remaining data, we also emphasize the importance of difficulty distribution and diversity, which significantly impact the model’s long CoT reasoning ability. We annotate a small-scale classification training dataset to construct a neural scorer that assigns a difficulty score to each document. Considering diversity, we do not entirely remove low-difficulty samples but apply appropriate proportional downsampling. For queries with multiple answers, we strive to balance answer sources or synthetic method and impose limits where the number of answers is excessively high.

Ultimately, we follow~{\longcatchat} to perform semantic similarity based data decontamination and duplication by combination of rule-based and nerual-based model.

\subsection{Details of SFT}
\label{appendix:sft}

\begin{figure}[b]
    \centering
    \includegraphics[width=0.35\linewidth]{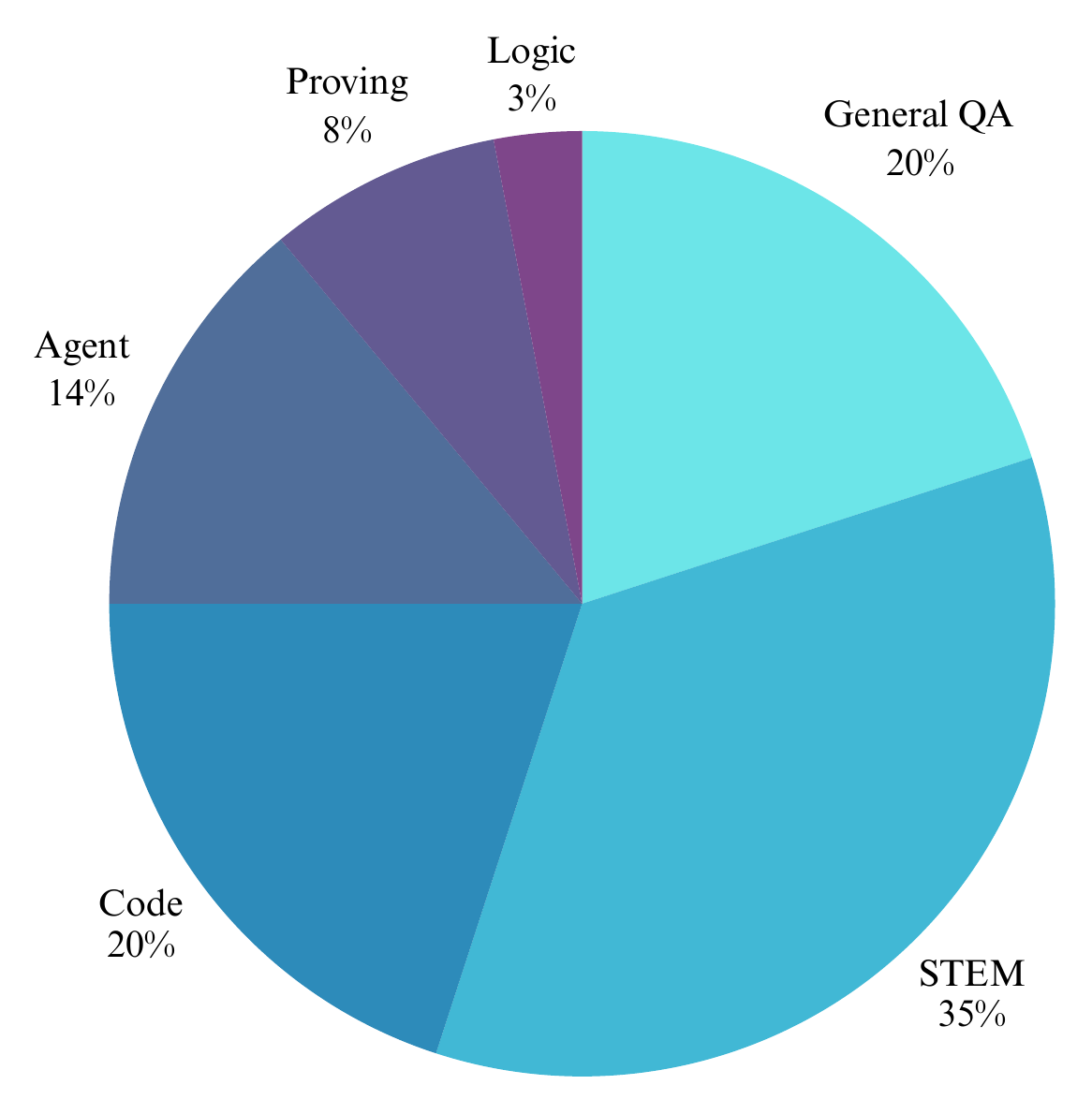}
    \caption{The distribution of our curated SFT data. 
    }
    \label{fig:sft_ratio}
\end{figure}

\subsubsection{Details of General Reasoning Data}

\noindent\paragraph{STEM} 
The training data of STEM contains several hundred thousand instances with verifiable answers,
spanning mathematics, physics, chemistry, biology, and other related science scenarios. 
The majority of questions are drawn from open-source datasets, public competitions, 
and a few instruction data retrieved from the pre-training corpus.

To ensure data quality and correctness, we implement a multi-stage filtering pipeline. In the initial phase, we utilize the LLM-as-a-Judge method to automatically identify and exclude questions that are not suitable for answering, such as incomplete statements, 
and conceptual queries.
Subsequently, for each remaining question, we employ several advanced LRMs to generate candidate responses.
Thus, we can obtain the voted results, and they can be used to filter the data whose ground truth is inconsistent with the voted results. 



\noindent\paragraph{Code}
The code data is mainly collected from open-source competition data. 
We screen out queries with the clear problem description, a set of unit tests with more than 5 test cases, and a judgment script with starter code or function body.
Specifically, the problem description must contain the question, running cost constraints, and some input-output pairs.
The unit test is viewed as the argument of the generated code and can be used to validate its correctness by the judgment script.

To avoid problems such as unclear title descriptions, incorrect unit tests, and judgment script errors, 
we first implement a filtering pipeline to eliminate queries containing garbled content, missing information, or logical errors.
We then build an in-house code sandbox environment and leverage multiple advanced LRMs to generate candidate codes. The data will be filtered that whose generated codes by all LRMs can not pass all unit tests in our sandbox. 
To balance diversity, we train a small classifier to tag the specific knowledge points and difficulty for each query, and perform difficulty filtering, ensuring that problems possess an appropriate level of complexity and applicability to real-world algorithmic reasoning.


\noindent\paragraph{Logic}
Logical reasoning
is one of the imperative general reasoning tasks and plays a significant role in human-like exploration, backtracking, and deep-thinking~\citep{yu2024natural, Jiang2025Do}. 
We gather a series of widely-considered logical tasks in terms of deductive, inductive, and abductive. 
For each task, we devise a specialized generator to synthesize queries with precisely controllable difficulty automatically. 
In that, we can categorize query difficulty into three distinct levels: easy, medium, and hard.

For synthesizing high-quality responses,
we first train a small model by employing RLVR~\citep{lambert2024tulu} on these logical tasks. 
Then, we generate multiple long CoT trajectories for each query by distilling from the trained model.
The reliable response with the correct answer will be used for the training instances.



\noindent\paragraph{General QA}
We also consider some general-purpose tasks, such as instruction-following, commonsense knowledge, and safety, etc. 
For instruction-following, we curate both single-turn and multi-turn instruction-following datasets, with varying levels of constraint complexity and quantity. 
We filter queries that have low semantic quality, and employ a reverse prompt generation strategy~\citep{femepid2024gradual} to guarantee the response satisfy all constraints.
For commonsense knowledge, we develop three types of general QA datasets: reading comprehension, table-based question answering, and custom-designed tasks. 
These specific domains can substantially enhance our model's multi-turn dialogue, reasoning, and long context abilities. 
For safety, we first develop a content safety policy that categorizes queries into more than 40 distinct safety categories across five response types: comply, comply with guideline, soft refuse, soft refuse with guideline, or hard refuse.
This criterion guides us to train a small model to divide each query into a specific safety category, and to optimize the response by human annotation based on different safety categories.
In addition, because over-refusal can significantly impact model helpfulness, we also optimize the refusal ability by carefully processing the general-purpose queries.


Except for general QA, furthermore, we assessed the difficulty of the query with the pass rate estimated by advanced LRMs.
Queries with a pass rate exceeding a predefined threshold were considered too easy and removed, promoting a focus on problems that require substantive reasoning. 
The final training set was sampled from the filtered pool based on the pass rate distribution, resulting in a high-quality dataset suitable for training reasoning models.


\end{document}